\documentclass{article}
\PassOptionsToPackage{numbers, compress}{natbib}
\usepackage[preprint]{icml2026}

\usepackage{microtype}
\usepackage{hyperref}
\usepackage{url}
\usepackage{booktabs}
\usepackage{graphicx}
\usepackage{algorithm}
\usepackage{algorithmic}
\usepackage{amssymb}
\usepackage{amsmath}
\usepackage{booktabs}
\usepackage{multirow}
\usepackage{wrapfig}
\usepackage{makecell}
\usepackage{pifont}

\usepackage{cleveref}
\usepackage{tabularx}
\usepackage{booktabs}
\usepackage{caption}
\usepackage[table]{xcolor}
\definecolor{rowblue}{RGB}{220,236,252}
\usepackage{tikz}
\definecolor{iconblue}{RGB}{31,119,180}
\definecolor{iconorange}{RGB}{255,127,14}
\definecolor{iconpurple}{RGB}{148,103,189}
\newcommand{\best}[1]{\textbf{#1}}
\newcommand{\second}[1]{\underline{#1}}
\newcommand{\spec}[1]{\textcolor{black!55}{#1}}

\definecolor{deltablue}{HTML}{BBDEFB}
\definecolor{deltaamber}{HTML}{FFF3CD}

\DeclareRobustCommand{\deltagood}[1]{%
  \tikz[baseline=(X.base)]{
    \node[fill=deltablue, rounded corners=1.5pt, inner xsep=1.2pt, inner ysep=0.2pt] (X)
    {\scriptsize #1};
  }%
}

\DeclareRobustCommand{\deltabad}[1]{%
  \tikz[baseline=(X.base)]{
    \node[fill=deltaamber, rounded corners=1.5pt, inner xsep=1.2pt, inner ysep=0.2pt] (X)
    {\scriptsize #1};
  }%
}

\DeclareRobustCommand{\goodbadge}{%
  \raisebox{0.15ex}{%
    \begingroup
    \setlength{\fboxsep}{1pt}%
    \colorbox{deltablue}{\rule{0pt}{1.0ex}\rule{1.6ex}{0pt}}%
    \endgroup
  }%
}

\DeclareRobustCommand{\badbadge}{%
  \raisebox{0.15ex}{%
    \begingroup
    \setlength{\fboxsep}{1pt}%
    \colorbox{deltaamber}{\rule{0pt}{1.0ex}\rule{1.6ex}{0pt}}%
    \endgroup
  }%
}

\makeatletter
\newcommand{\compactparagraph}[1]{%
  \par\noindent
  \refstepcounter{paragraph}%
  \addcontentsline{toc}{paragraph}{#1}%
  \textbf{#1}\hspace{0.5em}%
}
\makeatother

\newcounter{packednmbr}

\usepackage{amsthm}
\usepackage{xcolor}

\makeatletter
\DeclareFontFamily{OMS}{cmtt}{\skewchar\font48}
\DeclareFontShape{OMS}{cmtt}{m}{n}{<-> ssub * cmsy/m/n}{}
\DeclareFontShape{OMS}{cmtt}{m}{it}{<-> ssub * cmsy/m/n}{}
\DeclareFontShape{OMS}{cmtt}{m}{sl}{<-> ssub * cmsy/m/n}{}
\DeclareFontShape{OMS}{cmtt}{bx}{n}{<-> ssub * cmsy/b/n}{}
\DeclareFontShape{OMS}{cmtt}{b}{n}{<-> ssub * cmsy/b/n}{}
\makeatother

\definecolor{sveblue}{HTML}{1F77B4}
\definecolor{sveorange}{HTML}{D55E00}

\usepackage[most]{tcolorbox}

\definecolor{svepast}{HTML}{D55E00}
\definecolor{svenew}{HTML}{7A3E9D}

\newcommand{\past}[1]{\textcolor{svepast}{#1}}
\newcommand{\newparam}[1]{\textcolor{svenew}{#1}}

\newtheorem*{theorem-restate}{Theorem~\ref{thm:stateless-lb}}

\definecolor{citecolor}{HTML}{0b64c5}
\hypersetup{
 colorlinks=True,
 linkcolor=citecolor,
 citecolor=citecolor,
 urlcolor=citecolor}

\crefname{figure}{Fig.}{Figs.}
\crefname{figure}{Fig.}{Figs.}
\crefname{table}{Tab.}{Tabs.}
\crefname{table}{Tab.}{Tabs.}
\crefname{section}{\S\!\!}{\S\S\!\!}
\crefname{section}{\S\!\!}{\S\S\!\!}
\crefname{subsection}{\S\!\!}{\S\S\!\!}
\crefname{subsection}{\S\!\!}{\S\S\!\!}
\crefname{appendix}{Appx.}{Appx.}
\crefname{appendix}{Appx.}{Appx.}
\crefname{equation}{Eq.}{Eqs.}
\crefname{equation}{Eq.}{Eqs.}

\definecolor{boxgray}{RGB}{248,248,248}
\definecolor{boxblue}{RGB}{235,245,255}
\definecolor{boxgreen}{RGB}{238,248,240}
\definecolor{boxyellow}{RGB}{255,249,229}

\newtcolorbox{formatbox}[1]{
  enhanced,
  title=#1,
  colback=white,
  colframe=black,
  fonttitle=\bfseries,
  boxrule=0.6pt,
  arc=1.5pt,
  left=1.2mm,
  right=1.2mm,
  top=1mm,
  bottom=1mm,
  before skip=0.8em,
  after skip=0.8em
}

\newtcolorbox{promptbox}[1]{%
  breakable,
  enhanced,
  title={#1},
  colback=boxblue,
  colframe=black!70,
  fonttitle=\bfseries,
  boxrule=0.5pt,
  arc=1.5pt,
  left=1.2mm,
  right=1.2mm,
  top=1mm,
  bottom=1mm,
  before skip=0.6em,
  after skip=0.6em
}

\newtcolorbox{notebox}[1]{
  enhanced,
  title=#1,
  colback=boxyellow,
  colframe=black!65,
  fonttitle=\bfseries,
  boxrule=0.5pt,
  arc=1.5pt,
  left=1.2mm,
  right=1.2mm,
  top=1mm,
  bottom=1mm,
  before skip=0.6em,
  after skip=0.6em
}

\lstdefinestyle{jsonstyle}{
  basicstyle=\ttfamily,
  breaklines=true,
  columns=fullflexible,
  frame=single,
  backgroundcolor=\color{boxgray},
  xleftmargin=0pt,
  xrightmargin=0pt
}

\usepackage{silence}
\begin{document}

\twocolumn[
  \icmltitle{Stateful Visual Encoders for Vision-Language Models}

  \begin{icmlauthorlist}
    \icmlauthor{Zirui Wang}{voio,berkeley}
    \icmlauthor{Junwei Yu}{voio,berkeley}
    \icmlauthor{Adam Yala}{voio,berkeley,ucsf}
    \icmlauthor{David M. Chan}{berkeley}
    \icmlauthor{Joseph E. Gonzalez}{berkeley}
    \icmlauthor{Trevor Darrell}{voio,berkeley}
  \end{icmlauthorlist}

  \icmlaffiliation{voio}{Voio, Inc.}
  \icmlaffiliation{berkeley}{UC Berkeley}
  \icmlaffiliation{ucsf}{UCSF}

  \icmlcorrespondingauthor{Zirui Wang}{zwcolin@berkeley.edu}

  \icmlkeywords{Vision-Language Models, Visual Encoders, Multi-Image Reasoning}

  \vskip 0.3in
]

\printAffiliationsAndNotice{}

\vspace{-4ex}
\begin{abstract}
Vision-language models (VLMs) are increasingly used in multi-image, multi-turn agentic settings where decisions depend on visual changes.
However, in existing open-weight VLMs, visual comparisons happen only inside the language model, while the visual encoder itself remains \textit{stateless}: each image is encoded independently, without access to the prior visual context.
As a result, small but task-critical changes may be attenuated before the language model has a chance to compare them, especially when those changes do not affect the high-level semantics of the scene.
We introduce a \textbf{Stateful Visual Encoder }, which conditions each visual representation on prior visual features.
Under supervised finetuning, VLMs equipped with stateful encoders achieve consistent improvements on controlled tasks involving cross-image spatial aggregation, multi-object visual differencing, and visual trajectory behavior cloning.
These improvements are consistent across input resolutions, language model sizes, and VLM backbones.
Finally, we validate our model on real-world tasks, including longitudinal radiology, fine-grained image comparison, and remote sensing, where stateful consistently improve generalist VLM baselines and can match or surpass specialized models in selected domains.
Project page: \url{https://statefulvisualencoders.github.io/}
\vspace{-2ex}
\end{abstract}
\section{Introduction}
\begin{figure}
    \centering
    \includegraphics[width=\linewidth]{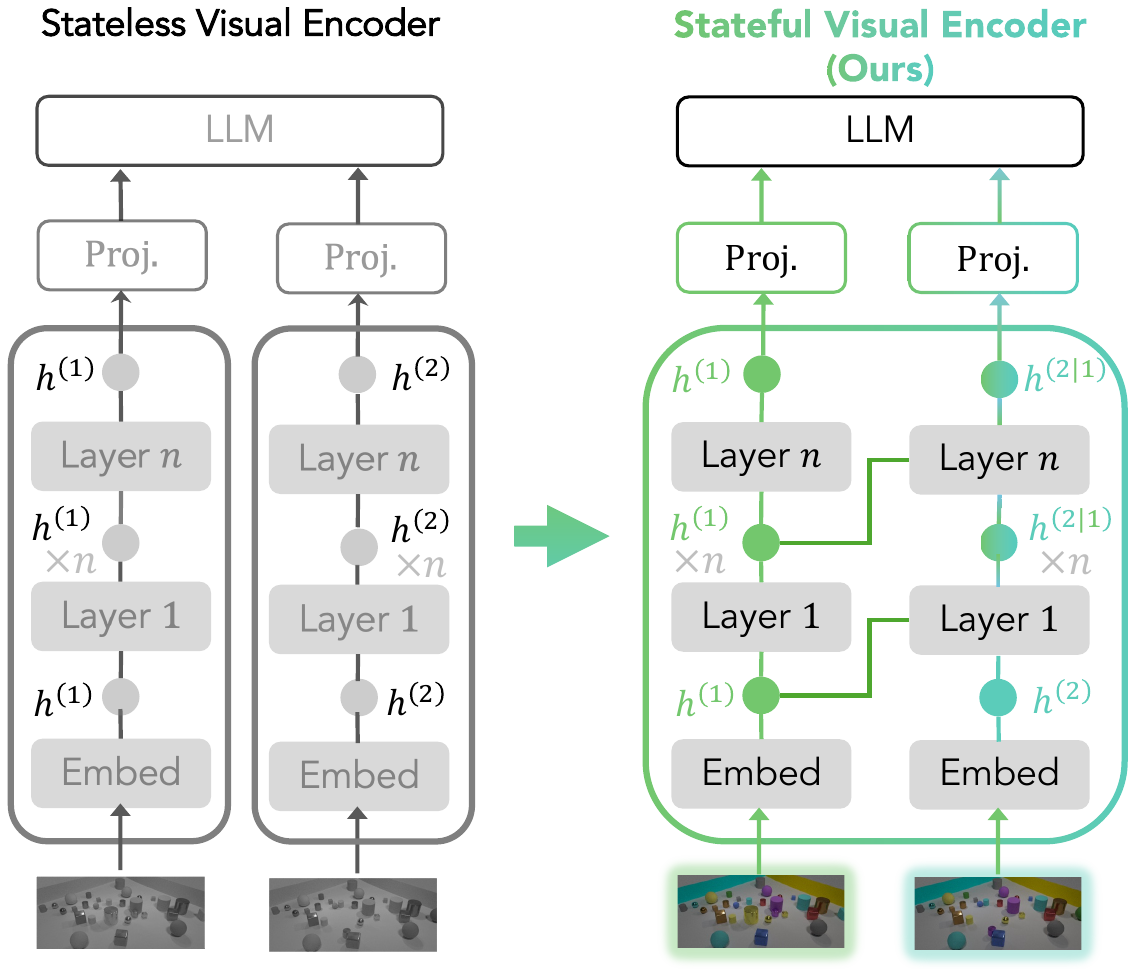}
    \vspace{-3ex}
\caption{
\small\textbf{Stateful visual encoders} condition each image's visual representation on features from the previous image within the vision backbone, enabling early cross-image comparison inside the visual encoder.
The left-to-right direction ensures that the current image can attend only to past visual features, matching interactions where future observations may not yet be available.
}
    \vspace{-4ex}
    \label{fig:teaser}
\end{figure}
Vision-language models (VLMs) are increasingly used in interactive and comparative visual tasks, where a model must observe, track, and analyze visual changes across images to make grounded decisions.
Despite such dynamic behavior, the dominant architecture of open-weight VLMs remains inherited from static image-language modeling: each image is passed independently through a visual encoder, and the resulting visual tokens are compared only later by the language model.
Thus, while the overall VLM may process a sequence of images, its visual encoder remains stateless.

This stateless encoding is limiting because visual changes are often subtle, for example, a chest X-ray finding may newly appear or partially resolve, a small structure may appear in a satellite image, or an edited image may differ only in a localized attribute.
These subtle changes are often critical to task performance.
Yet visual encoders used in modern VLMs~\citep{qwen3.5, bai2025qwen3, zeng2025glm, wang2025internvl35, Kamath2025Gemma3T} are typically pretrained for language-aligned~\citep{radford2021learning, zhai2023sigmoid} or self-supervised representations~\citep{caron2021emerging, tong2024eyes} and applied to each image independently.
As a result, per-image encoding can unintentionally suppress the fine-grained differences needed for comparison.

To address this, we add \textbf{cross-image interaction} (i.e., \cref{fig:teaser}) directly into the visual encoder, conditioning the current visual representation on features from previous images before passing tokens to the language model.
Using controlled synthetic tasks that require strict visual comparisons \citep{wang2025opencua, qiu2021describing, wang2026visgym}, we evaluate design choices for architecting (\cref{subsec:results}), initializing, and optimizing cross-image interactions (\cref{subsec:ablations}).
We study several lightweight variants, including extending self-attention context, adding cross-attention from current visual features to the prior features, augmenting this interaction with an FFN, and using adaptive normalization to condition visual features.
To preserve compatibility with pretrained VLMs, we initialize added interaction modules from nearby pretrained weights when possible, zero-initialize output branches to avoid disrupting the original feature distribution at the start of finetuning, and stop gradients through the prior features during cross-image retrieval.

We validate the effectiveness of SVEs both on synthetic domains, where we demonstrate that SVEs consistently improve task performance beyond what can be explained by simply adding parameters or compute, and on three real-world domains: detecting visual differences in radiology scans \citep{PhysioNet-medical-diff-vqa-1.0.1} (\cref{subsec:radiology}), performing fine-grained image comparison on edits derived from real-world/web images \citep{ye2025imgeditunifiedimageediting} (\cref{subsec:image_comparison}), and identifying changes in remote-sensing images \citep{9934924} (\cref{subsec:remote_sensing}).
Compared to naive finetuning, SVE-based encoders consistently improve model performance on these tasks and can match or surpass specialized models in selected domains.
Furthermore, these gains scale robustly across image resolutions ($256^2$--$768^2$), model sizes (0.8B--9B), and diverse VLM families, including Qwen3.5, Qwen3-VL, GLM-4.6V-Flash, InternVL3.5, and Gemma-3 (\cref{subsec:generality}).

Overall, our contributions can be summarized as follows:
(1) We introduce the \textbf{Stateful Visual Encoder (SVE)}, a simple architectural extension that injects cross-image interactions inside the visual encoder of open-weight VLMs without replacing the visual backbone or retraining the full model from scratch.
(2) We develop a practical SVE finetuning strategy, including initialization and optimization choices that stabilize finetuning and improve state-dependent visual representations in the SFT regime.
(3) We demonstrate the effectiveness and generality of SVEs across controlled visual comparison tasks, image resolutions, model sizes, and VLM families, and further validate it on real-world comparison tasks in radiology, image editing, and remote sensing.

\section{Related work}
\label{sec:related}

\begin{figure*}[t]
    \centering
    \includegraphics[width=\linewidth]{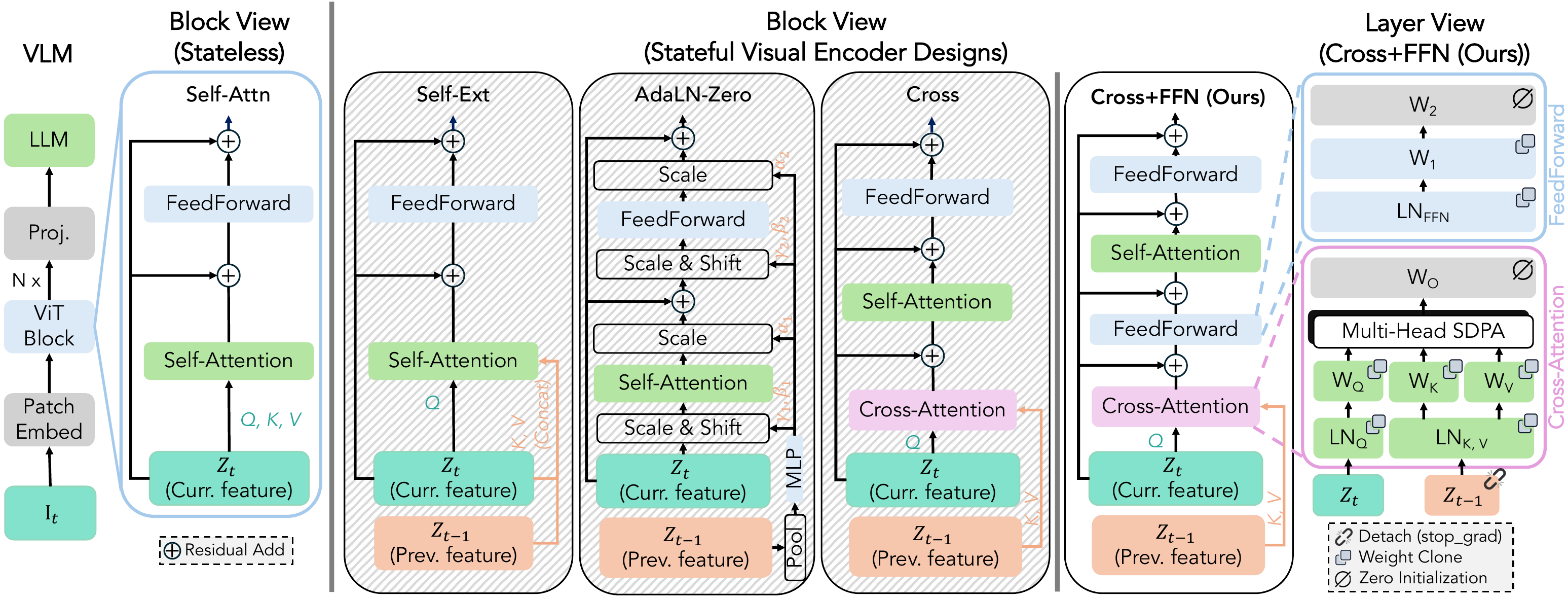}
    \caption{
    \small\textbf{Design study and implementation recipe for SVE}.
    We compare several ways to condition current visual tokens $Z_t$ on past tokens $Z_{t-1}$.
    The layer view expands the winning Cross-Attn + FFN design and shows its implementation recipe: stop-gradient on the past feature pathway, cloned initialization from the same ViT block, and zero initialization.
    Activations and positional embeddings in the layer view are omitted for simplicity.
    }
    \label{fig:design_and_recipe}
\end{figure*}

\compactparagraph{Image Difference Encoders.}
Specialized change-detection models compare images inside the visual encoder~\citep{park2019robust, chen2021bit, bandara2022changeformer, dong2024changeclip}.
However, unlike our SVE, these architectures are designed for specific change detection tasks, rather than studied as general-purpose visual encoders for VLMs.

\compactparagraph{Video Visual Encoders.}
Beyond pairwise change modeling, video encoders learn spatiotemporal representations from frame sequences.
Representative video encoders include I3D~\citep{carreira2017quo}, MViT~\citep{fan2021multiscale}, Video Swin~\citep{liu2022video}, TimeSformer~\citep{bertasius2021space}, and ViViT~\citep{arnab2021vivit}, with recent video foundation models and MLLMs such as VideoMAE~\citep{tong2022videomae}, InternVideo2~\citep{wang2024internvideo2}, and VideoPrism~\citep{zhao2024videoprism} scaling this direction through video-text supervision, masked modeling, and long-context spatiotemporal tokenization.
Recent video-aware encoders, such as Perception Encoder~\citep{bolya2026perception} and OneVision-Encoder~\citep{tang2026onevision}, further train visual backbones for both image and video understanding.
SVEs instead target \textit{image-based} VLMs that receive multiple images in context, such as sparse observations, before-after pairs, and interaction states.
Rather than training a spatiotemporal visual backbone, a SVE introduces causal cross-image conditioning into the existing image encoder: features of the current image condition on those from the previous image, while future images remain unavailable.
This matches interactive settings while preserving the existing VLM visual interface.

\compactparagraph{Multi-Image Encoding in VLMs.}
Recent VLMs~\citep{liu2023visual, liu2024improved, li2023blip, bai2025qwen3, dai2023instructblip, alayrac2022flamingo, bai2025qwen25vltechnicalreport} have shown strong multimodal reasoning abilities~\citep{qin2025chain, bigverdi2025perception}, but multi-image state reasoning remains challenging.
Most multi-image VLMs adopt late fusion:
methods such as MANTIS~\citep{jiang2024mantis}, LLaVA-NeXT-Interleave~\citep{li2024llavainterleave}, LLaVA-OneVision~\citep{li2024llavaonevision}, Idefics3~\citep{laurencon2024idefics3}, and VILA~\citep{lin2024vila} encode images independently and leave cross-image comparison to the language model.
Long-video and streaming VLMs add memory banks, token compression, or KV-cache mechanisms after visual encoding~\citep{he2024malmm, zhang2024flashvstream, diko2025rewind, shi2026attendattentionefficientscalable, xu2025streamingvlm}.
SVE addresses a complementary bottleneck: the current image can retrieve and integrate prior visual features inside the visual backbone before serialization to the LLM.

\section{Stateful Visual Encoders}
\label{sec:sve}

\compactparagraph{Background.}
Modern vision-language models (VLMs) typically consist of a visual encoder $f_V$, a vision-language connector $W$, and a large language model (LLM) $f_L$.
Given an image $I$ preprocessed into a sequence of $N$ image patches, the visual encoder maps patches into visual features $Z = f_V(I) \in \mathbb{R}^{N \times d_V}$, where $d_V$ is the hidden dimension of the visual encoder.
The connector $W_{Proj}$ then maps these visual features into the LLM embedding space as $H = W_{Proj}(Z) \in \mathbb{R}^{M \times d_L}$, where $d_L$ is the LLM hidden dimension and $M$ is the number of visual tokens passed to the LLM.

\compactparagraph{Overview.}
As shown in \Cref{fig:design_and_recipe}, we study four stateful encoder designs.
\textbf{Self-Ext} extends the pretrained self-attention key-value set with features from the previous image.
\textbf{AdaLN-Zero} pools features from the previous image to modulate the self-attention and feed-forward layers through adaptive normalization \citep{perez2018film,peebles2023scalable}.
\textbf{Cross} inserts a full token-level cross-attention layer before each pretrained self-attention layer, with queries from all visual tokens of the current image and keys/values from all visual tokens of the previous image.
\textbf{Cross+FFN} further adds a feed-forward block after the inserted cross-attention layer.
We summarize the block form, added parameters, and added compute of each design in \Cref{app:sve_different_view}.
We use controlled multi-image comparison tasks (\cref{subsec:task_setup}, \cref{subsec:train_setup}) to select the final design (\cref{subsec:results}).
We then ablate the recipe needed to exploit past visual features without destabilizing training (\cref{subsec:ablations}), and test generality across resolutions, model sizes, and model backbones (\cref{subsec:generality}).
Finally, we provide feature analysis on stateful visual representation in \cref{app:feature_analysis}, detailed evaluation protocol in \cref{app:eval_conventions} and training configurations in \cref{app:training_infrastructure}.

\subsection{Task Setup}
\label{subsec:task_setup}
\begin{figure*}[ht]
    \centering
    \includegraphics[width=\linewidth]{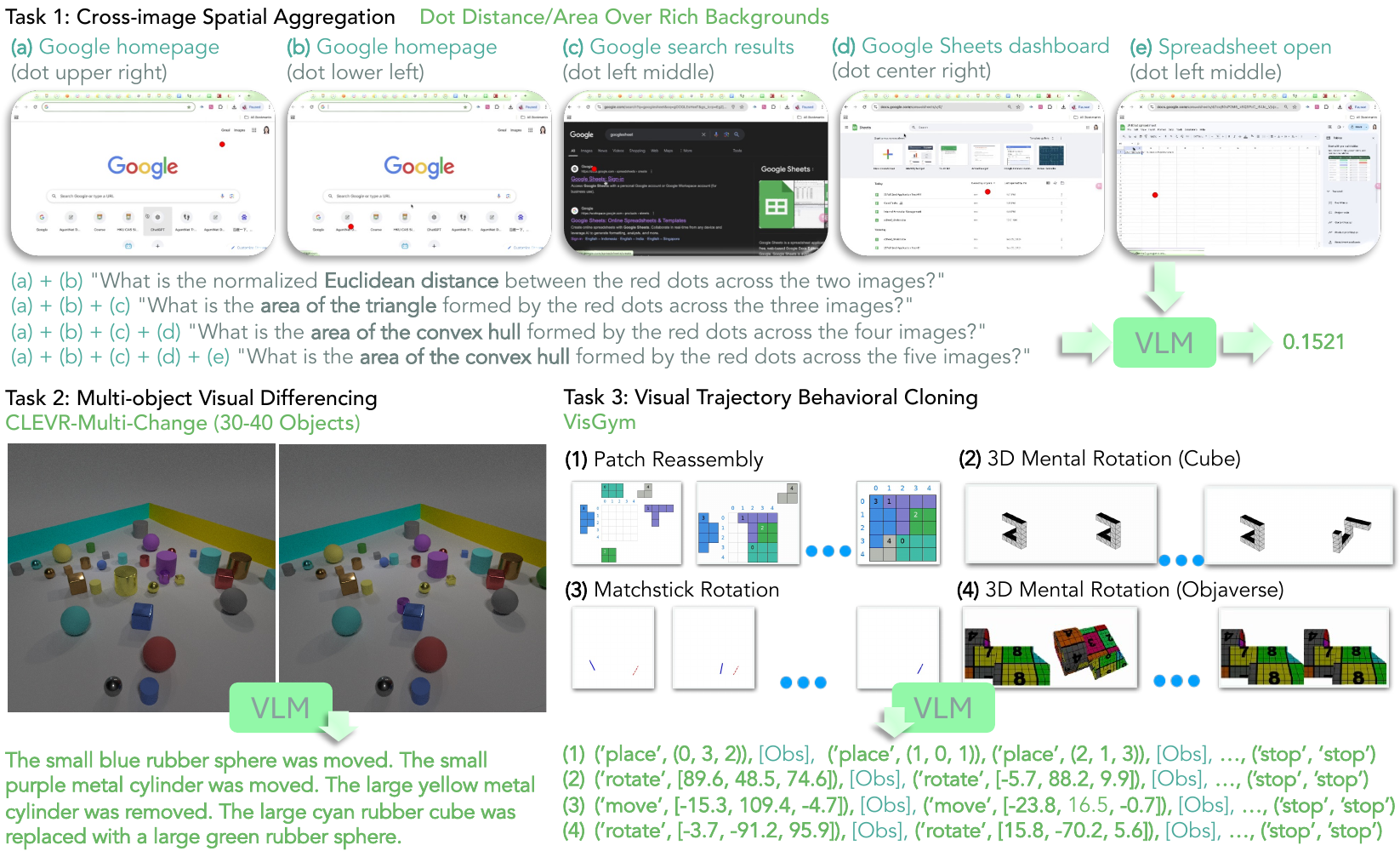}
    \caption{
    \small\textbf{Controlled tasks for studying stateful visual representations in vision-language models.}
    We present 3 tasks where we train and evaluate models with: cross-image spatial aggregation (top); multi-object visual differencing (bottom left); visual trajectory behavioral cloning (bottom right). Details are in \Cref{subsec:task_setup}.
    }
    \label{fig:controlled_tasks}
\end{figure*}
\compactparagraph{Cross-Image Spatial Aggregation.}
Image-text aligned visual encoders such as CLIP~\citep{radford2021learning} and SigLIP~\citep{zhai2023sigmoid} can struggle to expose fine-grained spatial or attribute information needed for downstream tasks~\citep{chen2024spatialvlm, pantazopoulos2024lost, bianchi2024clip}.
To isolate this failure mode in a controlled setting, we construct a spatial aggregation task that requires localizing small visual changes across semantically rich computer-use backgrounds from AgentNet/OpenCUA~\citep{wang2025opencua}.
We overlay random red dots across image sequences and ask the model to predict cross-image geometric quantities, including normalized Euclidean distance and convex-hull area (\Cref{fig:controlled_tasks}, top).
We report mean absolute error (MAE) and root mean square error (RMSE) on a held-out set.
Additional details on data formatting are available in \cref{app:spatial_aggregation_format}.

\compactparagraph{Multi-Object Visual Differencing.}
Spatial aggregation tests geometry but not whether a model can identify which object changed in a cluttered scene.
Using the CLEVR-Multi-Change engine~\citep{johnson2017clevr,qiu2021describing}, we create scene pairs with 30--40 objects and 4 subtle changes, including movement, insertion, deletion, and replacement (\Cref{fig:controlled_tasks}, bottom left).
The model must describe the changed object and change type.
We report exact-match accuracy for categorical change prediction, and BLEU~\citep{papineni-etal-2002-bleu}, CIDEr~\citep{vedantam2015cider}, METEOR~\citep{banerjee-lavie-2005-meteor}, SPICE~\citep{anderson2016spice}, and ROUGE-L~\citep{lin-2004-rouge} for generated descriptions.
Additional details on data formatting are available in \cref{app:clevr_multi_change_format}.

\compactparagraph{Visual Trajectory Behavioral Cloning.}
To test state tracking in interactive settings, we train models to imitate heuristic-solver demonstrations from VisGym~\citep{wang2026visgym}.
Each trajectory contains a task instruction followed by interleaved visual observations and solver actions, and the model predicts the next action from the interaction history.
We use four VisGym tasks: Patch Reassembly, 3D Mental Rotation (Cube), Matchstick Rotation, and 3D Mental Rotation (Objaverse), which require fine-grained perception, partial state tracking, and task-specific dynamics (\Cref{fig:controlled_tasks}, bottom right).
We report perplexity on a held-out set.
Additional details on data formatting are available in \cref{app:visgym_format}.

\subsection{Training Setup}
\label{subsec:train_setup}
\compactparagraph{Initialization.}
Unless otherwise specified, we initialize all new parameters inside the visual encoder of a pretrained Qwen3.5-4B~\citep{qwen3.5} model.
For each added cross-attention layer, we copy the input projections from the corresponding pretrained self-attention layer in the same visual-encoder block, i.e., $W_Q^{\mathrm{cross}}, W_K^{\mathrm{cross}}, W_V^{\mathrm{cross}} \leftarrow W_Q^{\mathrm{self}}, W_K^{\mathrm{self}}, W_V^{\mathrm{self}}$, and zero-initialize the output projection $W_O^{\mathrm{cross}}=\mathbf{0}$.
For the added FFN in Cross+FFN, we similarly copy the first linear layer and zero-initialize the second, i.e., $W_1^{\mathrm{cross}} \leftarrow W_1^{\mathrm{self}}$ and $W_2^{\mathrm{cross}}=\mathbf{0}$ (\Cref{fig:design_and_recipe}, right).
This gives the added modules a layer-matched feature basis while preserving the pretrained visual encoder's initial behavior \citep{kingma2018glow, bachlechner2021rezero, zhang2023adding}.

\compactparagraph{Conditioning.}
For cross-attention variants at each layer, the current visual features $Z_t$ provide queries and the predecessor visual features $Z_{t-1}$ provide keys and values: $Q_t = Z_t W_Q^{\mathrm{cross}}$ and $(K_{t}, V_{t})=(Z_{t-1}W_K^{\mathrm{cross}}, Z_{t-1}W_V^{\mathrm{cross}})$.
For the first image, we fall back to using $Z_1$ as the key-value source.
For Self-Ext, the key-value source is expanded from $Z_t$ to $[Z_t; Z_{t-1}]$.
For AdaLN-Zero, pooled predecessor visual features provide the conditioning vector, with a zero vector used for the first image.

\compactparagraph{Optimization.}
During training, we apply stop-gradient to the predecessor branch in all cross-attention variants reminiscent of BYOL and SimSiam~\citep{grill2020bootstrap,chen2021exploring}: $K_{t-1}=\texttt{stop\_grad}(Z_{t-1})W_K^{\mathrm{cross}}$ and $V_{t-1}=\texttt{stop\_grad}(Z_{t-1})W_V^{\mathrm{cross}}$ (\Cref{fig:design_and_recipe}, right). Gradients therefore update the current-image query branch and state-conditioning parameters, but not the features from the previous image used as context. We provide SFT hyperparameters in \Cref{tab:sft_hyperparams_all}.

\subsection{Results}
\label{subsec:results}
\begin{table*}[t]
\centering
\small
\caption{
\small \textbf{Cross-image spatial aggregation results}.
We report MAE/RMSE on dot-distance and area estimation tasks; all values are $\times 10^{-2}$.
Tri., Quad., and Pent. denote triangular, quadrilateral, and pentagon area estimation.
Colored badges show absolute change from the stateless baseline: \goodbadge{} indicates improvement and \badbadge{} indicates degradation.
}
\label{tab:spatial_aggregation_arch}
\setlength{\tabcolsep}{4.5pt}
\resizebox{\linewidth}{!}{
\begin{tabular}{
l
rr @{\hspace{6pt}}
rr @{\hspace{6pt}}
rr @{\hspace{6pt}}
rr @{\hspace{6pt}}
rr
}
\toprule
\multirow{3}{*}{Method}
& \multicolumn{2}{c}{Dot Distance}
& \multicolumn{2}{c}{Tri. Area}
& \multicolumn{2}{c}{Quad. Area}
& \multicolumn{2}{c}{Pent. Area}
& \multicolumn{2}{c}{Average} \\
&
\multicolumn{2}{c}{(2-Img)}
& \multicolumn{2}{c}{(3-Img)}
& \multicolumn{2}{c}{(4-Img)}
& \multicolumn{2}{c}{(5-Img)}
& \multicolumn{2}{c}{} \\
\cmidrule(lr){2-3}
\cmidrule(lr){4-5}
\cmidrule(lr){6-7}
\cmidrule(lr){8-9}
\cmidrule(lr){10-11}
&
MAE $\downarrow$ & RMSE $\downarrow$
& MAE $\downarrow$ & RMSE $\downarrow$
& MAE $\downarrow$ & RMSE $\downarrow$
& MAE $\downarrow$ & RMSE $\downarrow$
& MAE $\downarrow$ & RMSE $\downarrow$ \\
\midrule
Baseline (\textit{Stateless})
& 1.17 & 1.51
& 0.85 & 1.22
& 1.11 & 1.64
& 1.47 & 2.03
& 1.15 & 1.60 \\

\midrule

\textit{Stateful} \\
\quad Self-Ext.
& 1.55 \deltabad{.38} & 2.18 \deltabad{.67}
& 1.16 \deltabad{.31} & 1.72 \deltabad{.50}
& 1.35 \deltabad{.24} & 1.84 \deltabad{.20}
& 1.71 \deltabad{.24} & 2.35 \deltabad{.32}
& 1.44 \deltabad{.29} & 2.02 \deltabad{.42} \\

\quad AdaLN-Zero
& 1.23 \deltabad{.06} & 1.60 \deltabad{.09}
& 0.89 \deltabad{.04} & 1.26 \deltabad{.04}
& 1.12 \deltabad{.01} & 1.49 \deltagood{.15}
& 1.42 \deltagood{.05} & 2.05 \deltabad{.02}
& 1.17 \deltabad{.02} & 1.60 \deltabad{.00} \\

\quad Cross
& \second{0.97} \deltagood{.20} & \second{1.23} \deltagood{.28}
& \second{0.79} \deltagood{.06} & \second{1.15} \deltagood{.07}
& \second{1.03} \deltagood{.08} & \second{1.36} \deltagood{.28}
& \second{1.34} \deltagood{.13} & \second{1.84} \deltagood{.19}
& \second{1.03} \deltagood{.12} & \second{1.39} \deltagood{.21} \\

\rowcolor{rowblue}
\quad \textbf{Cross+FFN}
& \best{0.56} \deltagood{.61} & \best{0.72} \deltagood{.79}
& \best{0.50} \deltagood{.35} & \best{0.77} \deltagood{.45}
& \best{0.76} \deltagood{.35} & \best{1.02} \deltagood{.62}
& \best{1.04} \deltagood{.43} & \best{1.34} \deltagood{.69}
& \best{0.72} \deltagood{.43} & \best{0.96} \deltagood{.64} \\
\bottomrule
\end{tabular}}
\end{table*}

\begin{table*}[t]
\centering
\small
\caption{
\small \textbf{Results on visual differencing and trajectory behavioral cloning}.
For CLEVR, PPL, B4, C, M, S, R-L, and Acc denote perplexity, BLEU-4, CIDEr, METEOR, SPICE, ROUGE-L, and change accuracy.
For VisGym, MSR, PR, MRC, and MRO denote the Patch Reassembly, 3D Mental Rotation (Cube), Matchstick Rotation, and 3D Mental Rotation (Objaverse).
Colored badges show absolute change from the stateless baseline: \goodbadge{} indicates improvement and \badbadge{} indicates degradation.
}
\label{tab:visual_diff_state_tracking_arch}
\setlength{\tabcolsep}{2pt}
\resizebox{\linewidth}{!}{
\begin{tabular}{
l
rrrrrrr @{\hspace{8pt}}
rrrr
}
\toprule
\multirow{2}{*}{Method}
& \multicolumn{7}{c}{CLEVR-Multi-Change (30--40 Objects)}
& \multicolumn{4}{c}{VisGym (Perplexity)} \\
\cmidrule(lr){2-8}
\cmidrule(lr){9-12}
& PPL $\downarrow$ & B4 $\uparrow$ & C $\uparrow$ & M $\uparrow$ & S $\uparrow$ & R-L $\uparrow$ & Acc $\uparrow$
& MSR $\downarrow$ & PR $\downarrow$ & MRC $\downarrow$ & MRO $\downarrow$ \\
\midrule
Baseline (\textit{Stateless})
& 1.229
& 90.5
& 529.5
& 93.5
& 79.0
& 92.3
& 91.1
& 2.162
& 2.074
& 1.204
& \second{1.205} \\

\midrule
Self-Ext.
& 1.226 \deltagood{.003}
& \second{92.0} \deltagood{1.5}
& \second{538.1} \deltagood{8.6}
& \second{95.2} \deltagood{1.7}
& \second{80.0} \deltagood{1.0}
& \second{93.4} \deltagood{1.1}
& \second{92.5} \deltagood{1.4}
& 2.292 \deltabad{.130}
& 2.132 \deltabad{.058}
& 1.218 \deltabad{.014}
& 1.218 \deltabad{.013} \\

AdaLN-Zero
& 1.230 \deltabad{.001}
& 90.9 \deltagood{.40}
& 531.8 \deltagood{2.3}
& 93.8 \deltagood{.30}
& 79.1 \deltagood{.10}
& 92.4 \deltagood{.10}
& 91.4 \deltagood{.30}
& 2.152 \deltagood{.010}
& 2.069 \deltagood{.005}
& \second{1.201} \deltagood{.003}
& 1.207 \deltabad{.002} \\

Cross
& \second{1.225} \deltagood{.004}
& 88.5 \deltabad{2.0}
& 515.0 \deltabad{14.5}
& 91.5 \deltabad{2.0}
& 77.8 \deltabad{1.2}
& 90.2 \deltabad{2.1}
& 89.3 \deltabad{1.8}
& \second{2.145} \deltagood{.017}
& \second{2.009} \deltagood{.065}
& \second{1.201} \deltagood{.003}
& \second{1.205} \deltagood{.000} \\

\rowcolor{rowblue}
\textbf{Cross+FFN}
& \best{1.219} \deltagood{.010}
& \best{92.7} \deltagood{2.2}
& \best{543.9} \deltagood{14.4}
& \best{95.4} \deltagood{1.9}
& \best{80.1} \deltagood{1.1}
& \best{93.9} \deltagood{1.6}
& \best{92.7} \deltagood{1.6}
& \best{2.111} \deltagood{.051}
& \best{1.944} \deltagood{.130}
& \best{1.193} \deltagood{.011}
& \best{1.203} \deltagood{.002} \\
\bottomrule
\end{tabular}}
\end{table*}
We compare the stateless baseline with four SVE variants (\cref{fig:design_and_recipe}, left \& middle).

\compactparagraph{Cross-image spatial aggregation.}
\Cref{tab:spatial_aggregation_arch} shows that Cross+FFN performs best across all spatial aggregation tasks, with the largest gain occurring in Dot-Distance, suggesting that explicit state conditioning is especially useful for precise cross-image localization.
Self-Ext. performs \textit{worse} than the stateless baseline, suggesting that simply expanding the self-attention key-value set can disrupt the pretrained visual encoder.
AdaLN-Zero is more stable but remains close to the baseline, indicating that pooled feature conditioning from the previous image is too compressed for fine-grained spatial retrieval.
By contrast, Cross improves over the baseline, and Cross+FFN improves further, suggesting token-level retrieval and the added FFN both help transform cross-attended features before they are passed back into the visual block.

\compactparagraph{Multi-object visual differencing and visual trajectory behavioral cloning.}
\Cref{tab:visual_diff_state_tracking_arch} further validates this design choice on visual differencing (\textit{e.g.,} CLEVR-Multi-Change (30--40 objects)) and behavioral cloning (\textit{e.g.,} VisGym).
On CLEVR, Cross+FFN improves over the stateless baseline across perplexity, change accuracy, and all language-generation metrics, including CIDEr from $529.5$ to $543.9$ and accuracy from $91.1$ to $92.7$.
On VisGym, it also improves all four trajectory behavioral cloning tasks.
Other variants are less consistent or less effective.

\subsection{Ablations}
\label{subsec:ablations}
\begin{table*}[t]
\centering
\small
\caption{
\small \textbf{Spatial aggregation ablations}.
We ablate the Cross+FFN recipe and report MAE/RMSE; all values are $\times 10^{-2}$. Colored badges show absolute change from Cross+FFN: \goodbadge{} indicates improvement and \badbadge{} indicates degradation.
}
\setlength{\tabcolsep}{4pt}
\resizebox{\linewidth}{!}{
\begin{tabular}{
l
rr @{\hspace{8pt}}
rr @{\hspace{8pt}}
rr @{\hspace{8pt}}
rr @{\hspace{8pt}}
rr
}
\toprule
\multirow{2}{*}{Method}
& \multicolumn{2}{c}{Dot Dist.}
& \multicolumn{2}{c}{Tri. Area}
& \multicolumn{2}{c}{Quad. Area}
& \multicolumn{2}{c}{Pent. Area}
& \multicolumn{2}{c}{Average} \\
\cmidrule(lr){2-3}
\cmidrule(lr){4-5}
\cmidrule(lr){6-7}
\cmidrule(lr){8-9}
\cmidrule(lr){10-11}
& MAE $\downarrow$ & RMSE $\downarrow$
& MAE $\downarrow$ & RMSE $\downarrow$
& MAE $\downarrow$ & RMSE $\downarrow$
& MAE $\downarrow$ & RMSE $\downarrow$
& MAE $\downarrow$ & RMSE $\downarrow$ \\
\midrule
\rowcolor{rowblue} 
\textbf{Cross+FFN}
& \second{0.56} & \second{0.72}
& \best{0.50} & \best{0.77}
& \best{0.76} & \best{1.02}
& \best{1.04} & \best{1.34}
& \best{0.72} & \best{0.96} \\

\midrule
\multicolumn{11}{l}{\textit{Capacity-controlled baseline}} \\
\quad Self+FFN
& 0.62 \deltabad{.06} & 0.79 \deltabad{.07}
& 0.54 \deltabad{.04} & \second{0.80} \deltabad{.03}
& 0.84 \deltabad{.08} & 1.12 \deltabad{.10}
& \second{1.07} \deltabad{.03} & \second{1.42} \deltabad{.08}
& 0.77 \deltabad{.05} & 1.03 \deltabad{.07} \\

\midrule
\multicolumn{11}{l}{\textit{Ablations}} \\
\quad \textit{w/o} $W_{Q, K, V, 1}$ clone 
& \best{0.53} \deltagood{.03} & \best{0.71} \deltagood{.01}
& \second{0.52} \deltabad{.02} & 0.85 \deltabad{.08}
& \second{0.80} \deltabad{.04} & \second{1.05} \deltabad{.03}
& 1.12 \deltabad{.08} & 1.45 \deltabad{.11}
& \second{0.74} \deltabad{.02} & \second{1.02} \deltabad{.06} \\

\quad \textit{w/o} $W_{O, 2}$ zero-init
& 1.13 \deltabad{.57} & 1.49 \deltabad{.77}
& 0.85 \deltabad{.35} & 1.35 \deltabad{.58}
& 1.17 \deltabad{.41} & 1.57 \deltabad{.55}
& 1.56 \deltabad{.52} & 2.23 \deltabad{.89}
& 1.18 \deltabad{.46} & 1.66 \deltabad{.70} \\

\quad \textit{w/o} $Z_1$ $(K,V)$ fallback
& 0.64 \deltabad{.08} & 1.31 \deltabad{.59}
& 0.57 \deltabad{.07} & 0.86 \deltabad{.09}
& 0.81 \deltabad{.05} & 1.09 \deltabad{.07}
& 1.11 \deltabad{.07} & 1.49 \deltabad{.15}
& 0.78 \deltabad{.06} & 1.19 \deltabad{.23} \\

\quad \textit{w/o} \texttt{stop\_grad}$(K,V)$
& 0.64 \deltabad{.08} & 0.83 \deltabad{.11}
& 0.60 \deltabad{.10} & 0.92 \deltabad{.15}
& 0.89 \deltabad{.13} & 1.19 \deltabad{.17}
& 1.14 \deltabad{.10} & 1.54 \deltabad{.20}
& 0.82 \deltabad{.10} & 1.12 \deltabad{.16} \\

\quad \textit{w/o} pos-embed
& 0.58 \deltabad{.02} & 0.76 \deltabad{.04}
& 0.59 \deltabad{.09} & 0.90 \deltabad{.13}
& 0.89 \deltabad{.13} & 1.19 \deltabad{.17}
& 1.15 \deltabad{.11} & 1.50 \deltabad{.16}
& 0.80 \deltabad{.08} & 1.09 \deltabad{.13} \\

\midrule
Baseline (\textit{Stateless})
& 1.17 \deltabad{.61} & 1.51 \deltabad{.79}
& 0.85 \deltabad{.35} & 1.22 \deltabad{.45}
& 1.11 \deltabad{.35} & 1.64 \deltabad{.62}
& 1.47 \deltabad{.43} & 2.03 \deltabad{.69}
& 1.15 \deltabad{.43} & 1.60 \deltabad{.64} \\
\bottomrule
\label{tab:spatial_aggregation_abl}
\end{tabular}}
\end{table*}
\begin{table*}[t]
\centering
\small
\caption{
\small \textbf{Visual differencing and trajectory behavioral cloning ablations}.
We ablate the Cross+FFN recipe on CLEVR and VisGym.
For CLEVR, PPL, B4, C, M, S, R-L, and Acc denote perplexity, BLEU-4, CIDEr, METEOR, SPICE, ROUGE-L, and change accuracy.
For VisGym, MSR, PR, MRC, and MRO denote the Patch Reassembly, 3D Mental Rotation (Cube), Matchstick Rotation, and 3D Mental Rotation (Objaverse) tasks.
Bold/underline indicate best/second-best results.
Colored badges show absolute change from Cross+FFN: \goodbadge{} indicates improvement and \badbadge{} indicates degradation.
}
\setlength{\tabcolsep}{2pt}
\resizebox{\linewidth}{!}{
\begin{tabular}{
l
rrrrrrr @{\hspace{8pt}}
rrrr
}
\toprule
\multirow{2}{*}{Method}
& \multicolumn{7}{c}{CLEVR-Multi-Change  (30--40 Objects)}
& \multicolumn{4}{c}{VisGym} \\
\cmidrule(lr){2-8}
\cmidrule(lr){9-12}
& PPL $\downarrow$ & B4 $\uparrow$ & C $\uparrow$ & M $\uparrow$ & S $\uparrow$ & R-L $\uparrow$ & Acc $\uparrow$
& MSR $\downarrow$ & PR $\downarrow$ & MRC $\downarrow$ & MRO $\downarrow$ \\
\midrule
\rowcolor{rowblue} 
\textbf{Cross+FFN}
& \best{1.219}
& \second{92.7}
& \second{543.9}
& \best{95.4}
& \best{80.1}
& \second{93.9}
& \best{92.7}
& \best{2.111}
& 1.944
& \best{1.193}
& \best{1.203} \\

\midrule
\multicolumn{12}{l}{\textit{Capacity-controlled baseline}} \\
\quad Self+FFN
& 1.223 \deltabad{.004}
& 91.6 \deltabad{1.1}
& 537.2 \deltabad{6.7}
& 94.8 \deltabad{.60}
& 79.9 \deltabad{.20}
& 93.0 \deltabad{.90}
& 91.6 \deltabad{1.1}
& 2.126 \deltabad{.015}
& \second{1.938} \deltagood{.006}
& 1.198 \deltabad{.005}
& \second{1.204} \deltabad{.001} \\

\midrule
\multicolumn{12}{l}{\textit{Ablations}} \\
\quad \textit{w/o} $W_{Q, K, V, 1}$ clone 
& 1.223 \deltabad{.004}
& 92.2 \deltabad{.50}
& 538.7 \deltabad{5.2}
& 94.9 \deltabad{.50}
& 79.9 \deltabad{.20}
& 93.4 \deltabad{.50}
& 92.5 \deltabad{.20}
& 2.161 \deltabad{.050}
& \best{1.933} \deltagood{.011}
& 1.202 \deltabad{.009}
& 1.207 \deltabad{.004} \\

\quad \textit{w/o} $W_{O, 2}$ zero-init
& 1.238 \deltabad{.019}
& 91.0 \deltabad{1.7}
& 534.8 \deltabad{9.1}
& 94.2 \deltabad{1.2}
& 78.3 \deltabad{1.8}
& 92.8 \deltabad{1.1}
& 91.0 \deltabad{1.7}
& 2.319 \deltabad{.208}
& 2.636 \deltabad{.692}
& 1.221 \deltabad{.028}
& 1.220 \deltabad{.017} \\

\quad \textit{w/o} $Z_1$ $(K,V)$ fallback
& \second{1.221} \deltabad{.002}
& 92.2 \deltabad{.50}
& 541.0 \deltabad{2.9}
& \second{95.1} \deltabad{.30}
& \second{80.0} \deltabad{.10}
& 93.3 \deltabad{.60}
& 91.8 \deltabad{.90}
& 2.140 \deltabad{.029}
& 1.972 \deltabad{.028}
& \second{1.201} \deltabad{.008}
& 1.205 \deltabad{.002} \\

\quad \textit{w/o} \texttt{stop\_grad}$(K,V)$
& \best{1.219} \deltabad{.000}
& \best{93.0} \deltagood{.30}
& \best{544.4} \deltagood{.50}
& \best{95.4} \deltabad{.00}
& \best{80.1} \deltabad{.00}
& \best{94.0} \deltagood{.10}
& \second{92.6} \deltabad{.10}
& 2.143 \deltabad{.032}
& 1.943 \deltagood{.001}
& 1.203 \deltabad{.010}
& 1.205 \deltabad{.002} \\

\quad \textit{w/o} pos-embed
& 1.224 \deltabad{.005}
& 91.8 \deltabad{.90}
& 537.3 \deltabad{6.6}
& 94.5 \deltabad{.90}
& 79.5 \deltabad{.60}
& 93.2 \deltabad{.70}
& 92.0 \deltabad{.70}
& \second{2.112} \deltabad{.001}
& 1.947 \deltabad{.003}
& \second{1.201} \deltabad{.008}
& 1.207 \deltabad{.004} \\

\midrule
Baseline (\textit{Stateless})
& 1.229 \deltabad{.010}
& 90.5 \deltabad{2.2}
& 529.5 \deltabad{14.4}
& 93.5 \deltabad{1.9}
& 79.0 \deltabad{1.1}
& 92.3 \deltabad{1.6}
& 91.1 \deltabad{1.6}
& 2.162 \deltabad{.051}
& 2.074 \deltabad{.130}
& 1.204 \deltabad{.011}
& 1.205 \deltabad{.002} \\
\bottomrule
\label{tab:visual_diff_state_abl}
\end{tabular}}
\end{table*}

We ablate the main components of the Cross+FFN recipe in~\Cref{tab:spatial_aggregation_abl,tab:visual_diff_state_abl}.
Overall, Cross+FFN benefits from explicit cross-image access, $W_{Q,K,V,1}$ cloning, $W_{O,2}$ zero-initialization, the $H_1$ $(K,V)$ fallback, $\texttt{stop\_grad}(K,V)$, and positional embeddings in the cross-attention pathway.

\compactparagraph{Capacity-controlled baseline.}
Self+FFN uses the same added pathway as Cross+FFN but does \textit{not} attend to features from the previous image.
We use this to rule out the possibility that gains come merely from added parameters or FLOPs rather than statefulness.
Although it improves over the stateless baseline with the rest of our recipe, it remains below Cross+FFN on all tasks but patch reassembly in VisGym, the only task where visual comparison is not strictly required \citep{wang2026visgym}.

\compactparagraph{$W_{Q,K,V,1}$ clone.}
Removing $W_{Q,K,V,1}$ cloning gives generally weaker results, suggesting that copying the input-side cross-attention weights and the first FFN layer from its following self-attention block provides a useful layer-matched feature basis.

\compactparagraph{$W_{O,2}$ zero-init.}
Removing $W_{O,2}$ zero-initialization causes the \textit{largest} degradation.
This supports the role of zero initialization in preserving the pretrained encoder's feature distribution at the start of finetuning.
Without it, the newly added cross-attention and FFN branches can immediately perturb visual features in large magnitude before they enter the following pretrained self-attention and FFN layers, placing those layers off-distribution.

\compactparagraph{$Z_1$ $(K,V)$ fallback.}
Removing the $Z_1$ $(K,V)$ fallback replaces the first-image key-value source with a learned null embedding, suggesting the stateful pathway should attend to real visual features if possible.

\compactparagraph{\texttt{stop\_grad}$(K,V)$.}
Removing $\texttt{stop\_grad}(K,V)$ weakens spatial aggregation and gives mixed results on visual differencing.
This supports treating keys and values from previous image features as a stable retrieval context, rather than allowing them to co-adapt directly through the current image's cross-attention update.

\compactparagraph{pos-embed.}
Removing positional embeddings from the cross-attention degrades performance across the evaluated tasks, with especially large drops on spatial aggregation and visual differencing.
This suggests that preserving positional information in cross-image attention is important for state-dependent visual understanding.

\begin{figure*}[t]
    \centering
    \includegraphics[width=\linewidth]{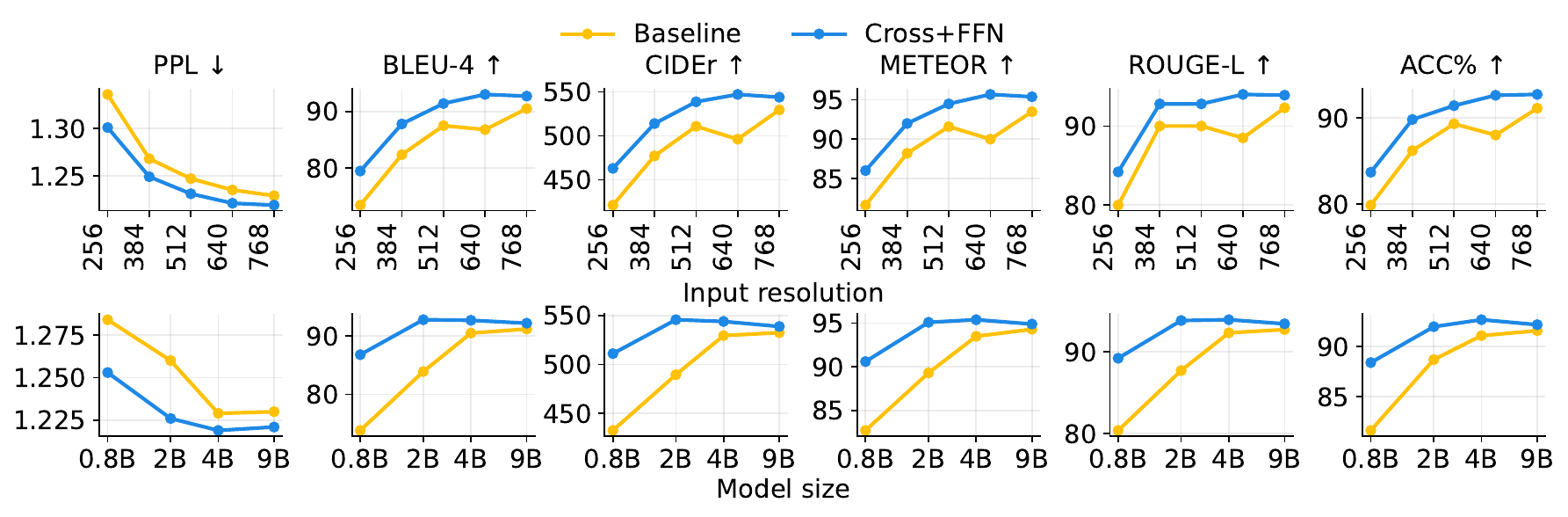}
\caption{
\small \textbf{SVE (Cross+FFN) generalizes across input resolutions and model sizes.}
We compare SVE (blue) with its stateless baseline (yellow) on multi-object visual differencing across input resolutions (top) and model sizes (bottom).
SVE consistently improves over the stateless baseline, especially when the baseline is weaker, while both approaches approach the task ceiling at higher resolutions and scales.
}
\label{fig:res_size_generalization}

\end{figure*}
\begin{table*}[t]
\centering
\small
\caption{
\small\textbf{SVE generalizes across different VLM backbones.}
We compare SVE with its stateless baseline on multi-object visual differencing across VLM backbones.
PPL, B4, C, M, S, R-L, and Acc denote perplexity, BLEU-4, CIDEr, METEOR, SPICE, ROUGE-L, and change accuracy.
\goodbadge{} indicates improvement over the corresponding stateless baseline.
}
\label{tab:backbone_generalization}
\setlength{\tabcolsep}{2.5pt}
\renewcommand{\arraystretch}{1.12}
\resizebox{\linewidth}{!}{
\begin{tabular}{@{}lllrrrrrrr@{}}
\toprule
\multirow{2}{*}{Backbone}
& \multicolumn{2}{c}{Backbone features}
& \multicolumn{7}{c}{CLEVR-Multi-Change (30--40 Objects)} \\
\cmidrule(lr){2-3}
\cmidrule(lr){4-10}
& Connector design & Distinct feature
& PPL $\downarrow$ & B4 $\uparrow$ & C $\uparrow$ & M $\uparrow$ & S $\uparrow$ & R-L $\uparrow$ & Acc $\uparrow$ \\
\midrule

Qwen3-VL-4B~\citep{bai2025qwen3}
& MLP merger $M=4/N$
& DeepStack \citep{meng2024deepstack}
& 1.268 \deltagood{.004}
& 82.5 \deltagood{2.5}
& 482.1 \deltagood{15.1}
& 88.6 \deltagood{1.3}
& 58.7 \deltagood{1.0}
& 86.8 \deltagood{1.5}
& 87.3 \deltagood{.70} \\

Qwen3.5-4B~\citep{qwen3.5}
& MLP merger $M=4/N$
& Gated DeltaNet \citep{yang2025gated}
& 1.219 \deltagood{.010}
& 92.7 \deltagood{2.2}
& 543.9 \deltagood{14.4}
& 95.4 \deltagood{1.9}
& 80.1 \deltagood{1.1}
& 93.9 \deltagood{1.6}
& 92.7 \deltagood{1.6} \\

GLM-4.6V-Flash~\citep{zeng2025glm}
& MLP merger $M=4/N$
& SwiGLU \citep{shazeer2020glu} FFN
& 1.236 \deltagood{.005}
& 92.4 \deltagood{.70}
& 542.0 \deltagood{3.8}
& 95.0 \deltagood{.40}
& 64.5 \deltagood{.40}
& 93.6 \deltagood{.40}
& 92.2 \deltagood{.10} \\

InternVL3.5-4B~\citep{wang2025internvl35}
& MLP merger $M=4/N$
& LayerScale \citep{touvron2021going}
& 1.332 \deltagood{.026}
& 68.2 \deltagood{1.7}
& 389.5 \deltagood{11.5}
& 77.8 \deltagood{1.1}
& 49.9 \deltagood{1.2}
& 76.3 \deltagood{1.2}
& 77.4 \deltagood{1.8} \\

Gemma-3-4B~\citep{Kamath2025Gemma3T}
& Pool to $M = 256 \quad \forall N$
& Local-global Attn. \citep{beltagy2020longformer}
& 1.316 \deltagood{.083}
& 68.4 \deltagood{8.0}
& 387.0 \deltagood{45.6}
& 78.0 \deltagood{5.9}
& 49.9 \deltagood{4.9}
& 76.3 \deltagood{5.9}
& 77.9 \deltagood{7.9} \\

\bottomrule
\end{tabular}}
\end{table*}

\subsection{Generality}
\label{subsec:generality}

We next evaluate the generality of SVEs (\textit{i.e., } the Cross+FFN recipe).
Specifically, we study whether the SVE design remains effective across different (1) input resolutions; (2) language model sizes; and (3) VLM backbones when compared to stateless baselines.
We use the \textit{multi-object visual differencing} task to train and evaluate all variants with two primary findings: \textbf{(1) SVEs are robust across input resolutions and model sizes.}
As shown in \Cref{fig:res_size_generalization}, SVEs consistently outperform a stateless baseline from $256^2$ to $768^2$ input resolution and from $0.8$B to $9$B model size.
Notably, smaller SVE models can \textit{match or even outperform} much larger stateless baselines.
\textbf{(2) SVEs generalize across VLM architectures.}
As shown in \Cref{tab:backbone_generalization}, SVEs consistently improve over stateless baselines across diverse VLM families, including Qwen3-VL~\citep{bai2025qwen3}, Qwen3.5~\citep{qwen3.5}, GLM-4.6V-Flash~\citep{zeng2025glm}, InternVL3.5~\citep{wang2025internvl35}, and Gemma-3~\citep{Kamath2025Gemma3T}.
These models differ substantially in visual encoders, vision--language connectors, attention mechanisms, and language backbones, suggesting that SVEs are not tied to a particular VLM architecture.

\section{Feature Analysis of Stateful Representations}
\label{app:feature_analysis}
\begin{figure*}[ht]
    \centering
    \includegraphics[width=\linewidth]{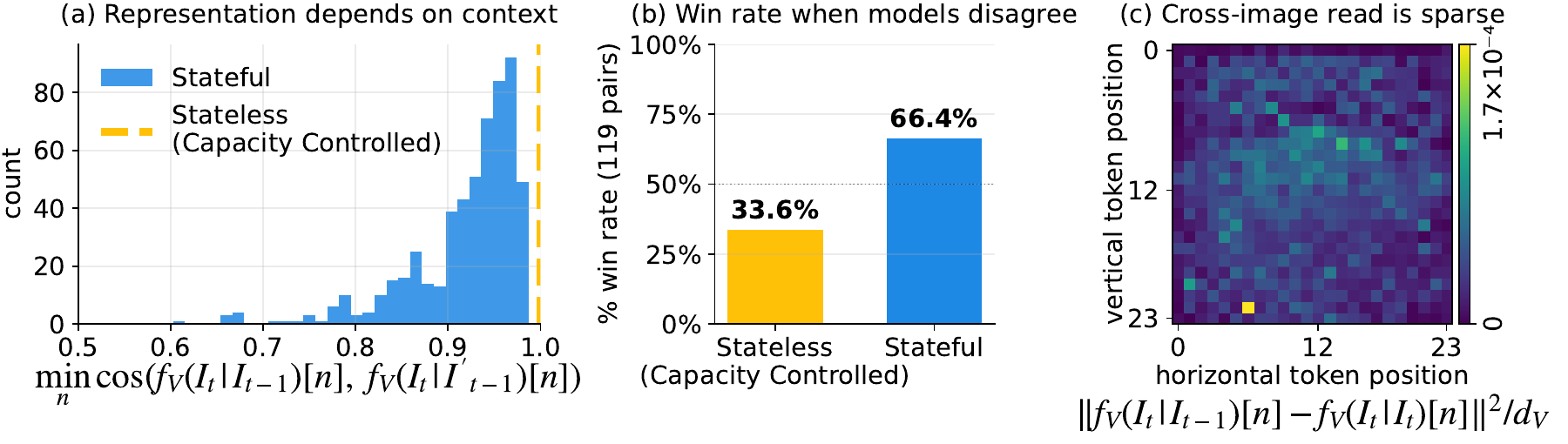}
    \vspace{-3ex}
\caption{
\textbf{Stateful encoding feature analysis.}
We compare SVE feature with the stateless baseline.
(a) SVE produces context-dependent visual features, while the stateless baseline remains unchanged.
(b) When the two models disagree, SVE wins the baseline by a large margin on CLEVR-Change.
(c) Cross-image feature updates are spatially sparse.
}
    \label{fig:feature_analysis}
\end{figure*}

We further analyze the learned visual features to understand state-dependent visual signals that lead to the gains of SVE.
We compare Cross+FFN against a capacity-controlled stateless baseline with the same architecture, trainable parameter count, training data, and optimization setup.
The only difference is the source of the temporal cross-attention keys and values: SVE reads from the features of the previous image, whereas the stateless control reads features from the current image itself, which is equivalently self-attention.
This comparison isolates whether the model learns to use past visual context, rather than merely benefiting from additional parameters or computation. 

Let
$
Z_t(Y) = f_V(I_t \mid Y) \in \mathbb{R}^{N \times d_V}
$
denote the visual representation of the current image $I_t$ when the state-conditioning source is $Y$, where $N$ is the number of spatial visual tokens and $d_V$ is the visual hidden dimension.
To measure context sensitivity, we compare the representation induced by the true predecessor (previous image) $I_{t-1}$ with the representation induced by a different predecessor $I'_{t-1}$:

\[
\resizebox{\linewidth}{!}{$
s_{\min}(I_t, I_{t-1}, I'_{t-1})
=
\min_{n \in \{1,\ldots,N\}}
\cos\!\left(
Z_t(I_{t-1})_n,
Z_t(I'_{t-1})_n
\right)
$}
\]

As shown in \cref{fig:feature_analysis}(a), the stateless control is invariant to predecessor swaps by construction, since there is no cross-image operation during visual encoding.
In contrast, SVE produces substantially lower minimum token similarity, indicating that the representation of $I_t$ depends on the preceding visual state with our introduced cross-image encoding module.

We next examine whether these context-dependent feature changes are useful for downstream change understanding.
Let $a_i^{\mathrm{sve}}$ and $a_i^{\mathrm{stateless}}$ denote the per-example Change-Acc scores of SVE and the stateless control on test example $i$.
We define the set of non-tied examples as
$
\mathcal{D}
=
\left\{
i : a_i^{\mathrm{sve}} \neq a_i^{\mathrm{stateless}}
\right\},
$
and compute the decided-example win rate
\[
\mathrm{WinRate}
=
\frac{1}{|\mathcal{D}|}
\sum_{i \in \mathcal{D}}
\mathbf{1}
\!\left[
a_i^{\mathrm{sve}} > a_i^{\mathrm{stateless}}
\right].
\]
As shown in \cref{fig:feature_analysis}(b), although many examples are ties due to the strength of the capacity-controlled baseline, SVE wins substantially more often among the non-tied cases.
This indicates that the state-dependent representation changes are not merely incidental feature shifts, but are predictive of improved visual change understanding.

Finally, \cref{fig:feature_analysis}(c) analyzes the spatial structure of the cross-image update.
For each test pair, we compare the SVE representation with the true predecessor against a masked-predecessor fallback, where the temporal cross-attention reads the current image itself:
\[
\Delta_n
=
\frac{1}{d_V}
\left\|
Z_t(I_{t-1})_n
-
Z_t(I_t)_n
\right\|_2^2 .
\]
We visualize the average $\Delta_n$ over test pairs on the post-merger spatial grid.
The resulting heatmap is sparse: most positions have near-zero update magnitude, while a small number of tokens absorb most of the cross-image change.
This supports the interpretation that SVE performs selective cross-image reading, preserving the pretrained visual representation for most tokens while updating localized features relevant to state comparison.
Together, these results show that SVE learns visual features that are context-dependent, task-relevant, and spatially selective.
\section{Validating SVE in Real-world Tasks}
\label{sec:real_world_tasks}
\begin{figure*}[t]
    \centering
    \includegraphics[width=\linewidth]{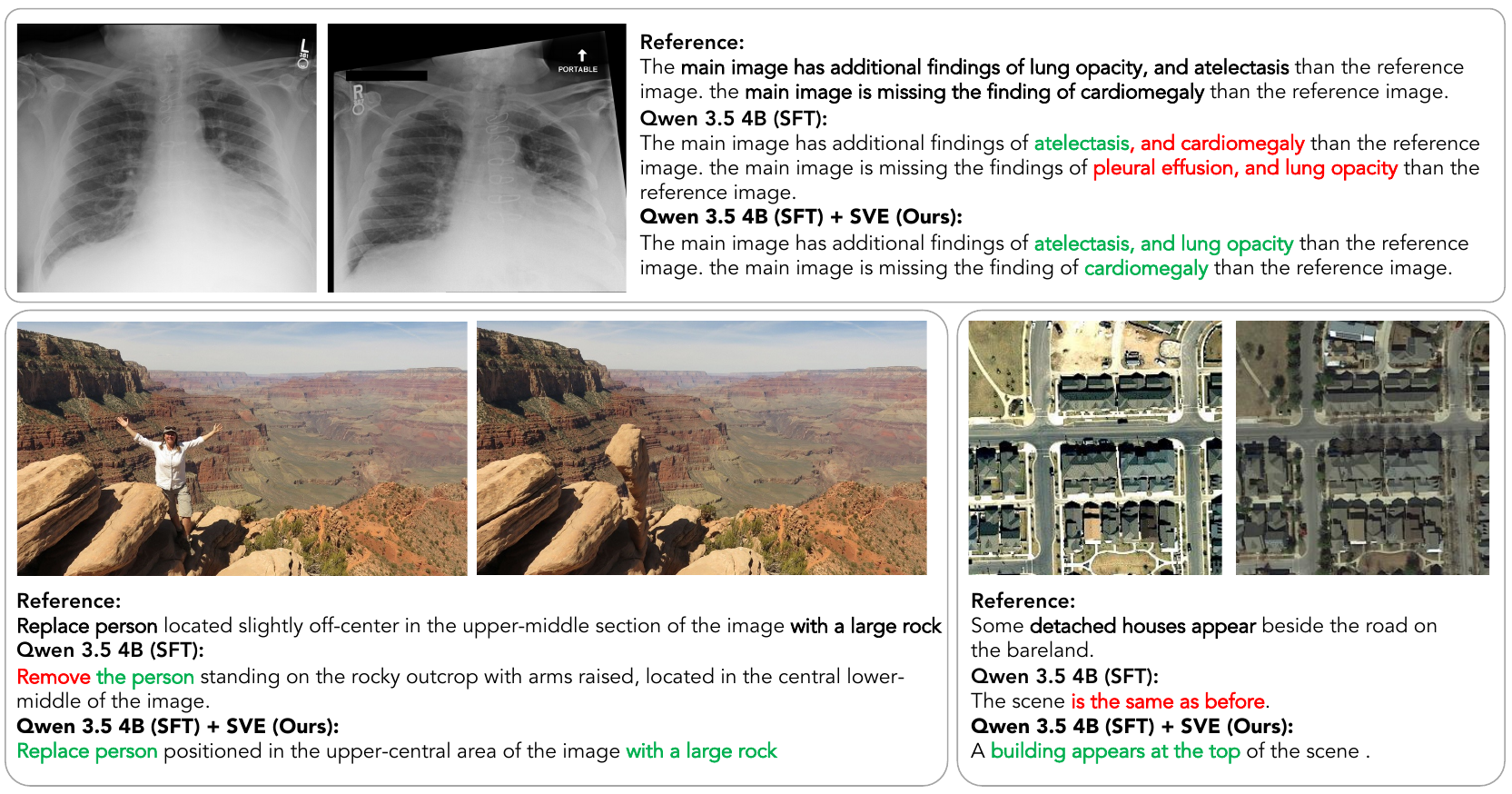}
\caption{
\small\textbf{Comparison of SVE vs. stateless baselines on real-world tasks.}
We show qualitative examples from longitudinal radiology (top), fine-grained image comparisons (bottom left), and remote sensing (bottom right).
Text in green and red indicates correct and incorrect change descriptions compared to the reference, respectively.
}
\vspace{-3ex}
    \label{fig:real_world_examples}
\end{figure*}

\begin{table}
\centering
\small
\caption{\small Medical-Diff-VQA evaluation results. We include captioning metrics i.e., BLEU-4 (B4), METEOR (M), ROUGE-L (R-L), and CIDEr (C) as well as evaluations based on RATE \citep{agrawal2025pillar}.}
\label{tab:radiology_performance}

\setlength{\tabcolsep}{4.5pt}
\resizebox{\linewidth}{!}{%
\begin{tabular}{lrrrrccc}
\toprule
\multirow{2}{*}{Method}
& \multirow{2}{*}{B4}
& \multirow{2}{*}{M}
& \multirow{2}{*}{R-L}
& \multirow{2}{*}{C}
& \multicolumn{2}{c}{Finding-level F1}
& \multirow{2}{*}{Change Acc.} \\
\cmidrule(lr){6-7}
& & & & & Micro & Macro & \\
\midrule
Qwen3.5 4B (SFT)
& 47.9 & 40.6 & 62.7 & 145.1
& 31.55 & 11.95 & 86.83 \\
\rowcolor{rowblue}
\quad +SVE (Ours)
& \best{49.6} & \best{40.9} & \best{66.3} & \best{178.9}
& \best{32.20} & \best{12.45} & \best{89.21} \\
\bottomrule
\end{tabular}
}
\end{table}

We validate the effectiveness of SVEs and our training recipe on three real-world comparison settings: detecting visual differences in radiology scans \citep{PhysioNet-medical-diff-vqa-1.0.1} (\cref{subsec:radiology}), performing fine-grained image comparison on edits derived from real-world/web images \citep{ye2025imgeditunifiedimageediting} (\cref{subsec:image_comparison}), and identifying changes in remote-sensing images \citep{9934924} (\cref{subsec:remote_sensing}). We provide additional details on data formatting in \cref{app:diffvqa_format}, \cref{app:imgedit_format}, \cref{app:levircc_format}, training configurations in \cref{app:training_infrastructure}, and evaluation setup in \cref{app:eval_conventions}.

\subsection{Longitudinal Radiology}
\label{subsec:radiology}

We first validate SVEs in longitudinal radiology, where clinically meaningful diagnostics often require fine-grained comparison across time.
We use the Medical-Diff-VQA dataset~\citep{PhysioNet-medical-diff-vqa-1.0.1}, which provides 16{,}347 paired chest X-ray images from the same patient together with annotations describing medical changes between the two studies. A SVE enables a VLM to better capture subtle longitudinal changes and therefore provides more grounded diagnostics (\Cref{fig:real_world_examples}, top), and achieves gains in standard captioning metrics (\cref{tab:radiology_performance}, left).

We further introduce a structured evaluation based on the RATE framework~\citep{agrawal2025pillar} to measure whether models capture clinically meaningful changes across 27 chest-related finding types (\textit{e.g.}, lung opacity, pneumothorax, calcification) in \cref{tab:radiology_performance} (right). We evaluate each X-ray pair by comparing the model’s predicted checklist of added or resolved findings against the reference, then report Micro/Macro F1 and whether it correctly detected any clinical change (additional details in \cref{app:diffvqa_eval}).
SVEs outperform the stateless baseline across all three metrics.

\subsection{Fine-grained Image Comparison}
\label{subsec:image_comparison}
\begin{table}
\centering
\small
\caption{
\small\textbf{ImgEdit evaluation results} under MLLM-as-a-judge.
We report pairwise preference counts against the baseline and reference instruction.
}
\label{tab:image_comparison_performance}
\vspace{-1ex}

\setlength{\tabcolsep}{5.5pt}
\renewcommand{\arraystretch}{1.05}
\resizebox{\linewidth}{!}{
\begin{tabular}{lrrr}
\toprule
Baseline
& Base Win
& Tied
& \cellcolor{rowblue}SVE Win \\
\midrule
Reference
& 296
& 758
& \cellcolor{rowblue}\best{346} \\
Qwen3.5 4B (SFT)
& 171
& 1020
& \cellcolor{rowblue}\best{209} \\
\bottomrule
\end{tabular}}

\end{table}
To test whether a SVE enables better image comparison in VLMs on real-world web images, we use ImgEdit~\citep{ye2025imgeditunifiedimageediting}, which consists of source images, edited images, and edit instructions.
Given a source--edited image pair, the model predicts the instruction that transformed the source image into the edited image.
This setting is directly relevant to edit verification~\citep{ma2024i2ebench}, image-editing reward modeling~\citep{luo2025editscore, wu2025editreward}, and image-difference understanding~\citep{baraldi2025changed,li2025superedit,di2025difftell}, all of which require models to compare before--after images and reason about whether the observed visual difference matches, explains, or refines a requested change (\cref{fig:real_world_examples}, bottom-left).

We train and evaluate SVEs against the stateless baseline on a subset of seven change categories (200 images each): add, adjust, background change, content memory, content understanding, remove, and replace~\citep{ye2026imgedit}.
We exclude categories where shortcut solutions exist, such as style change, where the style may reveal the target instruction without requiring comparison. Results are in \cref{tab:image_comparison_performance}.

Here, we opt out of traditional reference-based metrics because the reference edit instruction is not guaranteed to match the actual transformation.
Instead, we evaluate the output using a strong MLLM judge (Claude-Opus-4.7 \citep{anthropic2026claudeopus47}), and report pairwise preferences for SVEs against both the stateless baseline and the original reference instruction, where the SVE is preferred over both.

\begin{table}
\centering
\scriptsize
\caption{
\small\textbf{LEVIR-CC evaluation results} in comparison with prior methods.
$S_m^*$~\citep{10271701} averages BLEU-4 (B4), METEOR (M), ROUGE-L (R-L), and CIDEr (C).
}
\label{tab:remote_sensing_performance}
\vspace{-1ex}

\setlength{\tabcolsep}{2.5pt}
\renewcommand{\arraystretch}{1.0}
\resizebox{\linewidth}{!}{%
\begin{tabular}{lrrrr|r}
\toprule
Method & B4 & M & R-L & C & $S_m^*$ \\
\midrule
\multicolumn{6}{l}{\textit{Specialist models \& architectures}} \\
\spec{Capt-Diff~\citep{park2019robust}}
& \spec{47.41} & \spec{34.47} & \spec{65.64} & \spec{110.57} & \spec{64.52} \\
\spec{Capt-Rep~\citep{park2019robust}}
& \spec{53.15} & \spec{36.58} & \spec{69.73} & \spec{121.22} & \spec{70.17} \\
\spec{Capt-Att-Dual-Att~\citep{park2019robust}}
& \spec{53.56} & \spec{37.16} & \spec{69.19} & \spec{124.42} & \spec{71.08} \\
\spec{DUDA~\citep{park2019robust}}
& \spec{57.79} & \spec{37.15} & \spec{71.04} & \spec{124.32} & \spec{72.58} \\
\spec{MCCFormer-S~\citep{qiu2021describing}}
& \spec{56.36} & \spec{39.60} & \spec{69.46} & \spec{120.39} & \spec{71.45} \\
\spec{MCCFormer-D~\citep{qiu2021describing}}
& \spec{56.38} & \spec{39.91} & \spec{70.44} & \spec{124.44} & \spec{72.79} \\
\spec{RSICCFormer-C~\citep{9934924}}
& \spec{62.41} & \spec{38.70} & \spec{73.60} & \spec{132.62} & \spec{76.83} \\
\spec{PSNet~\citep{liu2023progressive}}
& \spec{62.11} & \spec{38.80} & \spec{73.60} & \spec{132.62} & \spec{76.78} \\
\spec{Chg2Cap~\citep{chang2023changes}}
& \spec{62.98} & \spec{39.42} & \spec{74.34} & \spec{136.25} & \spec{78.25} \\
\spec{SEN~\citep{10530145}}
& \spec{64.09} & \spec{39.59} & \spec{71.50} & \spec{125.02} & \spec{75.05} \\
\spec{Diffusion-RSCC~\citep{yu2025diffusion}}
& \spec{60.90} & \spec{37.80} & \spec{71.50} & \spec{125.60} & \spec{73.95} \\
\spec{CTMTNet~\citep{10740028}}
& \spec{64.69} & \spec{39.49} & \spec{74.54} & \spec{134.94} & \spec{78.42} \\
\spec{PromptCC~\citep{10271701}}
& \spec{63.54} & \spec{38.82} & \spec{73.72} & \spec{136.44} & \spec{78.13} \\
\spec{SAGE-CC~\citep{wang2025sam}}
& \spec{65.50} & \spec{39.92} & \spec{74.77} & \spec{137.50} & \spec{79.42} \\
\spec{SACNet~\citep{yang2026spatial}}
& \spec{65.57} & \spec{40.30} & \spec{75.68} & \spec{138.34} & \spec{79.97} \\
\midrule
\multicolumn{6}{l}{\textit{Generalist VLMs}} \\
Qwen3.5 4B (SFT)
& 60.70 & 39.42 & 76.03 & 142.26 & 79.60 \\
\rowcolor{rowblue}
\quad +SVE (Ours)
& \best{61.33} & \best{39.91} & \best{76.26} & \best{144.35} & \best{80.46} \\
\bottomrule
\end{tabular}}

\end{table}

\subsection{Remote Sensing}
\label{subsec:remote_sensing}

Remote sensing change captioning requires models to compare before-after aerial or satellite images of the same geographic region and describe how the scene has changed in the later image, such as newly constructed buildings, removed infrastructure, or altered land use (\cref{fig:real_world_examples}, bottom-right).
This task is a natural fit for SVEs because the task-relevant signal often lies in small, localized differences between the two images, while the surrounding geographic context remains largely unchanged.
To this end, we train and evaluate SVEs on LEVIR-CC~\citep{9934924}, a standard remote sensing change captioning dataset. We use standard captioning metrics and $S_m^*$ following prior work~\citep{10271701}, and present results in \cref{tab:remote_sensing_performance}. SVEs improve over the stateless baseline, and
moreover, SVEs outperforms all prior specialist models and architectures.

\section{Conclusion}

We presented the Stateful Visual Encoder (SVE), a simple yet effective method for introducing cross-image interaction into the visual encoder of a VLM.
SVEs consistently outperform stateless baselines across both synthetic datasets and real-world applications, from longitudinal radiology to remote sensing, and scales robustly across resolutions, model sizes, and architectures.
Overall, our results show that making the visual encoder state-aware can substantially improve multi-image reasoning while preserving the pretrained VLM interface, offering a practical path toward VLMs that better track, compare, and reason over dynamic visual contexts.

\paragraph{Acknowledgements}
We thank Kate Saenko, Mayank Mishra, Sanjay Sriram Subramanian, Kumar Krishna Agrawal, Lisa Dunlap, Natalia Harguindeguy, Baifeng Shi, XuDong Wang and Fangzhou Zhao for their discussion and/or support in developing this project.
Authors, as part of their affiliation with UC Berkeley, were supported by gifts from Accenture, AMD, Anyscale, Broadcom, Cisco, Google, IBM, Intel, Intesa Sanpaolo, Lambda, Lightspeed, Mibura, Microsoft, NVIDIA, Qualcomm, Samsung SDS, and SAP.

\bibliographystyle{plainnat}
\bibliography{main}

\newpage
\appendix

\renewcommand{\topfraction}{1.0}
\renewcommand{\textfraction}{0.00}
\renewcommand{\floatpagefraction}{1.0}
\raggedbottom

\clearpage
\newpage

The appendix is organized as follows:
\begin{itemize}
    \item \autoref{app:limitations} discusses some limitations of our approach.
    \item \autoref{app:training_data_formatting} describes the training data formatting for all six task families used in our experiments, including controlled synthetic tasks and real-world comparison tasks.
    \item \autoref{app:eval_conventions} defines evaluation-time prompt construction and the metric conventions used across tasks.
    \item \autoref{app:training_infrastructure} reports the software environment, distributed-training setup, tokenized cache, task-specific hyperparameters, and hardware infrastructure.
    \item \autoref{app:sve_different_view} provides a table view of the SVE design space and the per-layer parameter and compute overhead of each variant.
    \item \autoref{app:diffvqa_eval} provides additional details for the finding-level Medical-Diff-VQA evaluation protocol.
    \item \autoref{app:ai} discusses AI use in the preparation of this work.
\end{itemize}

\section{Limitations}
\label{app:limitations}

Although a SVE improves multi-image reasoning across controlled and real-world comparison tasks, several limitations remain.

\compactparagraph{Boundary of visual comparison.}
Our current formulation conditions each image on its immediate previous image at each visual-encoder layer. This still allows information to propagate \textit{diagonally} across images over depth, so the effective cross-image receptive field can grow with the number of layers.
However, long-range evidence is accessed only indirectly through intermediate visual states, rather than by explicit attention to all prior images. This is suitable for before--after comparison and short visual trajectories, but may be insufficient when relevant evidence is distributed across many earlier observations.

\compactparagraph{Domains that benefit from capturing changes.}
Our real-world evaluations focus on image-pair or image-sequence comparison in radiology, image editing, and remote sensing.
These domains cover diverse visual changes, but they do not fully capture the broader range of multimodal state tracking required in embodied agents, robotics, tactile interaction, audio-visual perception, or long-running computer-use environments.

\compactparagraph{Computational overhead.}
A SVE introduces additional cross-image computation inside the visual encoder (\cref{tab:sve_design_cost}). This overhead is usually modest compared to scaling the language backbone, but it can become nontrivial as image resolution, sequence length, or the number of visual states increases. Scaling stateful visual encoding to very long visual histories will therefore require more efficient memory, retrieval, or sparse attention mechanisms.

\section{Training Data Formatting}
\label{app:training_data_formatting}

This appendix documents the training data format for the six task families used to derive, train and evaluate SVE (\cref{sec:sve}, \cref{sec:real_world_tasks}):

\begin{enumerate}
    \item \textbf{Cross-image Spatial Aggregation}
    (\cref{subsec:task_setup}; Dot Distance/Area Over Rich Backgrounds~\citep{wang2025opencua});
    \item \textbf{Multi-object Visual Differencing}
    (\cref{subsec:task_setup}; CLEVR-Multi-Change (30--40 Objects)~\citep{johnson2017clevr,qiu2021describing});
    \item \textbf{Visual Trajectory Behavioral Cloning}
    (\cref{subsec:task_setup}; VisGym~\citep{wang2026visgym});
    \item \textbf{Longitudinal Radiology}
    (\cref{subsec:radiology}; Medical-Diff-VQA~\citep{PhysioNet-medical-diff-vqa-1.0.1});
    \item \textbf{Fine-grained Image Comparison}
    (\cref{subsec:image_comparison}; ImgEdit~\citep{ye2025imgeditunifiedimageediting});
    \item \textbf{Remote Sensing}
    (\cref{subsec:remote_sensing}; LEVIR-CC~\citep{9934924}).
\end{enumerate}

For each task, we describe the data source, conversation structure, number of images, system prompt, filler turns, supervision masking, and task-specific features. 
We use LlamaFactory (LF) \citep{zheng2024llamafactory} as the underlying infrastructure for all experiments, and Transformers backbone \citep{wolf-etal-2020-transformers} for inference and evaluation.
To ensure consistency and reproducibility, all experiments used seed 42 for data preparation, so the same data sequence applies to every model we train for any single task.

\begin{formatbox}{Common formatting conventions}
\begin{itemize}
    \item \textbf{Schema.} All non-VisGym datasets are pre-converted to LF's ShareGPT-style JSONL. Each line contains \texttt{messages} and \texttt{images}. The \texttt{k}-th \texttt{<image>} tag binds to \texttt{images[k]}.
    \item \textbf{VisGym exception.} VisGym uses the older ShareGPT conversation schema with \texttt{\{"from": "human"/"gpt", "value": ...\}} and embeds image paths inside conversation.
    \item \textbf{Image placement.} Each \verb|<image>| tag appears at the start of a user message, followed by a newline and task text.
    \item \textbf{History mask.} With \texttt{mask\_history=True}, only the final answer is supervised for single-shot datasets. Intermediate filler responses are masked out.
    \item \textbf{Template.} All tasks use LF's \texttt{qwen3\_5} chat template with \texttt{enable\_thinking=false}.
\end{itemize}
\end{formatbox}

\subsection{Cross-image Spatial Aggregation}
\label{app:spatial_aggregation_format}
\vspace{-4ex}
\begin{formatbox}{Cross-image Spatial Aggregation}
We use Dot Distance/Area Over Rich Backgrounds, where backghrounds are sampled from AgentNet ~\citep{wang2025opencua}.
This is a synthetic visual-numeric task where the model estimates a distance or area from red-dot locations across multiple screenshots. We downsample backgrounds to $384\times216$ for efficient experimentations.
\end{formatbox}

\begin{formatbox}{Conversation format}
\begin{lstlisting}[style=jsonstyle]
[system]    <task-specific system prompt>
[user]      <image>
            A red dot is placed on this screenshot.
[assistant] I see the red dot on the screenshot.
[user]      <image>
            A red dot is placed on this screenshot.
[assistant] I see the red dot on the screenshot.
...
[user]      <image>
            A red dot is placed on this screenshot.
            What is the {distance|area} formed by the red dots
            across the {two|three|four|five} images?
[assistant] 0.2555
\end{lstlisting}

\textbf{mask\_history:} \texttt{True}. Only the final numeric answer is supervised. The filler response \texttt{I see the red dot on the screenshot.} is masked.
\end{formatbox}

\begin{table}[ht]
\centering
\footnotesize
\setlength{\tabcolsep}{3pt}
\renewcommand{\arraystretch}{1.06}
\caption{Cross-image spatial aggregation summary.}
\label{tab:spatial_aggregation_tasks}
\begin{tabular*}{\linewidth}{@{\extracolsep{\fill}} l c p{0.31\linewidth} r r @{}}
\toprule
Sub-task & Img. & Quantity & Train & Eval \\
\midrule
\texttt{dot\_dist.} & 2 & Norm. Euclidean dist. & 100k & 1k \\
\texttt{tri\_area} & 3 & Norm. triangle area & 100k & 1k \\
\texttt{quad\_area} & 4 & Norm. convex-hull area & 100k & 1k \\
\texttt{pent\_area} & 5 & Norm. convex-hull area & 100k & 1k \\
\midrule
\textbf{Total} & -- & -- & \textbf{400k} & \textbf{4k} \\
\bottomrule
\end{tabular*}
\end{table}

\clearpage
\begin{promptbox}{System prompts}
\small
\begin{description}
    \item[2 images.]
    {\ttfamily\sloppy
    You are a visual distance estimator. You are shown two screenshots, each with a red dot.
    Your task is to estimate the normalized Euclidean distance between the red dots across the two images.
    The distance is normalized to [0, 1] where 0 means the dots are at the same position and 1 means
    they are at opposite corners. Output only the distance as a decimal number rounded to 4 decimal places.
    \par}

    \item[3 images.]
    {\ttfamily\sloppy
    You are a visual area estimator. You are shown three screenshots, each with a red dot.
    Your task is to estimate the normalized area of the triangle formed by the red dots across the three images.
    The area is normalized by the full image area, so it ranges from 0 to 0.5.
    Output only the area as a decimal number rounded to 4 decimal places.
    \par}
    
\end{description}
\end{promptbox}

\begin{promptbox}{System prompts}
\small
\begin{description}
    \item[4 images.]
    {\ttfamily\sloppy
    You are a visual area estimator. You are shown four screenshots, each with a red dot.
    Your task is to estimate the area of the convex hull formed by the red dots across the four images.
    The area is normalized by the full image area.
    Output only the area as a decimal number rounded to 4 decimal places.
    \par}

    \item[5 images.]
    {\ttfamily\sloppy
    You are a visual area estimator. You are shown five screenshots, each with a red dot.
    Your task is to estimate the area of the convex hull formed by the red dots across the five images.
    The area is normalized by the full image area.
    Output only the area as a decimal number rounded to 4 decimal places.
    \par}
\end{description}
\end{promptbox}

\subsection{Multi-object Visual Differencing}
\label{app:clevr_multi_change_format}

\begin{formatbox}{Multi-object Visual Differencing}
We use CLEVR-Multi-Change (30--40 Objects)~\citep{johnson2017clevr,qiu2021describing}, a two-image change-captioning task where the model describes 4 simultaneous object edits between a before and after CLEVR scene. The original CLEVR-Multi-Change dataset only involves 3--10 objects, which appear to be overly simple for current VLMs. We instead use its underlying engine to generate a dataset involving 30--40 objects with a wider camera angle under $768 \times 768$ resolution. There is a total of 100,000 training samples and 1,000 test samples.
\end{formatbox}

\begin{formatbox}{Conversation format}
\begin{lstlisting}[style=jsonstyle]
[user]      <image>
            Here is an image of a scene with objects.
[assistant] I see the scene. Please show me the next image.
[user]      <image>
            What changed between the two images?
[assistant] A new large yellow metal cylinder is visible.
            The large cyan rubber cube gave up its position to a
            large cyan metal cylinder. Someone replaced the large
            red rubber cylinder with a large yellow metal sphere.
            A new small red rubber cube is visible.
\end{lstlisting}

\textbf{mask\_history:} \texttt{True}. Only the final change description is supervised.
\end{formatbox}

\subsection{Visual Trajectory Behavioral Cloning}
\label{app:visgym_format}

\begin{notebox}{Schema difference}
VisGym uses the older ShareGPT-conversations schema:
\texttt{\{"from": "human"/"gpt", "value": "..."\}}.
The task description is placed in the first human turn, and each later human turn contains environment feedback plus the new visual observation.
\end{notebox}

\begin{formatbox}{Visual Trajectory Behavioral Cloning}
We use VisGym~\citep{wang2026visgym}, an episodic multi-turn visual reasoning task. Each human turn provides an updated image observation, and each GPT turn is a real action rather than a filler response. We curate SFT data from oracle solver demonstrations that is available in \url{https://huggingface.co/datasets/VisGym/visgym_data}. 
Each turn contains an observation of which the image resolution ranges from $336 \times 336$ to $256 \times 128$, depending on the task.
\end{formatbox}

\begin{table}[t]
\centering
\scriptsize
\setlength{\tabcolsep}{2.5pt}
\renewcommand{\arraystretch}{1.08}
\caption{Visual trajectory imitation task summary.}
\label{tab:visgym_task_summary}
\begin{tabular*}{\linewidth}{@{\extracolsep{\fill}} l r p{0.34\linewidth} @{}}
\toprule
Sub-task & Train & Description \\
\midrule
\texttt{matchstick\_rotation}
& 100k
& Move and rotate a blue stick to match a red target stick. \\

\texttt{mental\_rotation\_3d\_cube}
& 100k
& Rotate a colored cube to match a target orientation. \\

\begin{tabular}[t]{@{}l@{}}
\texttt{mental\_rotation}\\
\texttt{\_3d\_objaverse}
\end{tabular}
& 100k
& Rotate an Objaverse object to match a target view. \\

\texttt{patch\_reassembly}
& 100k
& Place irregular pieces to fill a $6{\times}6$ board. \\

\midrule
\textbf{Combined train}
& \textbf{400k}
& Union of four sub-tasks. \\

\textbf{Combined eval}
& \textbf{4k}
& 1k examples per sub-task. \\
\bottomrule
\end{tabular*}
\end{table}

\begin{formatbox}{Conversation format}
\begin{lstlisting}[style=jsonstyle]
[human] <image>
        {task description with action format}
        This is step 1. You are allowed to take K more steps.
[gpt]   <think>{reasoning}</think>
        <answer>('place', (0, 2, 4))</answer>

[human] <image>
        Environment feedback: Action executed successfully.
        This is step 2. You are allowed to take K-1 more steps.
[gpt]   <think>{reasoning}</think>
        <answer>('place', (1, 0, 1))</answer>

...
[human] <image>
        Environment feedback: ...
        This is step T. You are allowed to take 0 more steps.
[gpt]   <think>{reasoning}</think>
        <answer>('stop', 'stop')</answer>
\end{lstlisting}

\textbf{mask\_history:} \texttt{False}. All turns are supervised because every turn is a real action instead of a filler response.
\end{formatbox}

\subsection{Longitudinal Radiology}
\label{app:diffvqa_format}

\begin{formatbox}{Task summary: Longitudinal Radiology}
We use Medical-Diff-VQA~\citep{PhysioNet-medical-diff-vqa-1.0.1}, a medical change-captioning task where the model compares a reference chest X-ray and a current chest X-ray from the same patient. 
We convert raw \texttt{.dicom} files to high resolution (e.g., $3056 \times 2544$) \texttt{.png} files.
During training, we downsize images to $\sim 768^2$ due to memory overhead.
\end{formatbox}

\begin{table}[t]
\centering
\small
\setlength{\tabcolsep}{5pt}
\renewcommand{\arraystretch}{1.08}
\caption{Medical-Diff-VQA dataset summary.}
\label{tab:medical_diff_vqa_summary}
\begin{tabular}{lr}
\toprule
Property & Value \\
\midrule
Train samples & 130,335 \\
Validation samples & 12,573 \\
Test samples & 16,347 \\
Images per sample & 2 \\
\bottomrule
\end{tabular}
\end{table}

\begin{promptbox}{System prompt}
\small
{\ttfamily\sloppy
You are a medical imaging expert. Given a reference chest X-ray and a current chest X-ray from the same patient, describe what findings have changed.
\par}
\end{promptbox}

\begin{formatbox}{Conversation format}
\begin{lstlisting}[style=jsonstyle]
[system]    You are a medical imaging expert. Given a reference
            chest X-ray and a current chest X-ray from the same
            patient, describe what findings have changed.
[user]      <image>
            This is the reference (prior) chest X-ray.
[assistant] Understood. Please provide the current chest X-ray.
[user]      <image>
            This is the current chest X-ray.
            What has changed compared to the reference image?
[assistant] the main image has an additional finding of pneumothorax
            than the reference image. the main image is missing the
            findings of fracture, lung opacity, and pleural effusion
            than the reference image.
\end{lstlisting}

\textbf{Target format:} mostly templated finding-list descriptions, including no-change, additional findings, missing findings, or both.

\textbf{mask\_history:} \texttt{True}. Only the final diagnostic answer is supervised.
\end{formatbox}

\subsection{Fine-grained Image Comparison}
\label{app:imgedit_format}

\begin{formatbox}{Task summary: Fine-grained Image Comparison}
We use a subset of ImgEdit~\citep{ye2025imgeditunifiedimageediting} and repurpose it as a free-form image-pair or image-sequence comparison task. The model describes what changed between the original image and edited version, or across a sequence of edits. During training, we downsize images to $\sim 384^2$ due to memory overhead.
\end{formatbox}

\begin{table}[ht]
\centering
\small
\setlength{\tabcolsep}{5pt}
\renewcommand{\arraystretch}{1.08}
\caption{ImgEdit format summary.}
\label{tab:imgedit_v2_format}
\begin{tabular}{lr}
\toprule
Property & Value \\
\midrule
Train samples & 301,142 \\
Test samples & 1,400 \\
Images/sample & 2 or 3 \\
System prompts & 3 variants \\
\bottomrule
\end{tabular}
\end{table}

\begin{promptbox}{System prompt variants (depending on change category)}
\small
\begin{itemize}
    \item \texttt{You are given a pair of images. Describe the edit made between them.}
    \item \texttt{You are given a sequence of image edits. Describe all the changes made.}
    \item \texttt{You are given a sequence of image edits. Describe all the changes made and identify the underlying editing rule.}
\end{itemize}
\end{promptbox}

\begin{formatbox}{Pair-format conversation}
\begin{lstlisting}[style=jsonstyle]
[system]    You are given a pair of images. Describe the edit made
            between them.
[user]      <image>
            Here is the original image.
[assistant] I see the image. Please show me the edited version.
[user]      <image>
            What was edited between the two images?
[assistant] Turn bread positioned in the central to lower right area
            into toasted
\end{lstlisting}
\end{formatbox}

\begin{formatbox}{Sequence-format conversation}
\begin{lstlisting}[style=jsonstyle]
[system]    You are given a sequence of image edits. Describe all
            the changes made and identify the underlying editing rule.
[user]      <image>
            Here is the original image.
[assistant] I see the image. Please show me the next version.
[user]      <image>
            Here is the next version.
[assistant] I see the image. Please show me the next version.
[user]      <image>
            What was edited across all these versions?
[assistant] All subsequent edits must incorporate a warm, earthy tone
            adjust the house's color and texture adjust the brick patio
            area's color and material
\end{lstlisting}

\textbf{mask\_history:} \texttt{True}. Only the final edit description is supervised.
\end{formatbox}

\begin{notebox}{Known dataset issue}
Reference captions are derived from the instruction given to the editing model, not always from the visual edit that actually occurred. When the editing model fails to follow the instruction, the reference may describe a change that is not visually present.
\end{notebox}

\subsection{Remote Sensing}
\label{app:levircc_format}

\begin{formatbox}{Task summary: Remote Sensing}
We use LEVIR-CC~\citep{9934924}, a bitemporal remote-sensing change-captioning task where the model compares before and after satellite images of the same location.
Images in this dataset have a fixed resolution of $256 \times 256$.
\end{formatbox}

\begin{table}[t]
\centering
\small
\setlength{\tabcolsep}{5pt}
\renewcommand{\arraystretch}{1.08}
\caption{LEVIR-CC dataset summary.}
\label{tab:levir_cc_summary}
\begin{tabular}{lr}
\toprule
Property & Value \\
\midrule
Train captions & 34,075 \\
Validation captions & 6,665 \\
Full test captions & 1,929 \\
Images per sample & 2 \\
\bottomrule
\end{tabular}
\end{table}

\begin{promptbox}{System prompt}
\small
{\ttfamily\sloppy
You are an expert at detecting and describing changes between two images. Given a before and after image, describe what has changed.
\par}
\end{promptbox}

\begin{formatbox}{Conversation format}
\begin{lstlisting}[style=jsonstyle]
[system]    You are an expert at detecting and describing changes
            between two images. Given a before and after image,
            describe what has changed.
[user]      <image>
            This is the before satellite image.
[assistant] Understood. Please provide the after image.
[user]      <image>
            This is the after satellite image. Describe what has changed.
[assistant] there is no difference .
\end{lstlisting}

\textbf{Target format:} free-form change captions. Roughly half of training captions are no-change variants, while the rest describe changes.
\end{formatbox}

\section{Evaluation metric conventions}
\label{app:eval_conventions}
At evaluation time, prompts are constructed to end immediately before the final assistant turn. Therefore, filler text contributes to training context but is not generated as part of the test-time answer. All test sets are held out from training, with no overlap in image-pair and target-answer keys.
\subsection{Metric protocol by task}

Each task uses an evaluation protocol matched to its output type and to the dominant convention in prior published work on the corresponding benchmark. Numeric tasks are evaluated as regression (as discrete text tokens), agentic tasks are evaluated through action likelihood, and captioning tasks are evaluated with standard image-captioning metrics. Details are in \cref{tab:metric_protocol_by_task}.

\begin{table*}[ht]
\centering
\small
\setlength{\tabcolsep}{4pt}
\renewcommand{\arraystretch}{1.12}
\begin{tabularx}{\linewidth}{
>{\raggedright\arraybackslash}p{0.24\linewidth}
>{\raggedright\arraybackslash}p{0.30\linewidth}
>{\raggedright\arraybackslash}X
}
\toprule
Task & Reported metrics & Protocol \\
\midrule
\textbf{Cross-image Spatial Aggregation}
& MAE, RMSE
& Numeric regression over parsed decimal outputs. We report errors for dot-distance and area-estimation subtasks, with all values scaled by $10^{-2}$. \\

\textbf{Multi-object Visual Differencing}
& PPL, B4, C, M, S, R-L, Acc
& Permutation-invariant per-change captioning. We report perplexity, BLEU-4, CIDEr, METEOR, SPICE, ROUGE-L, and change accuracy. \\

\textbf{Visual Trajectory Behavioral Cloning}
& MSR, PR, MRC, MRO
& Agentic imitation evaluated by per-task perplexity. MSR, PR, MRC, and MRO denote Patch Reassembly, 3D Mental Rotation (Cube), Matchstick Rotation, and 3D Mental Rotation (Objaverse). \\

\textbf{Longitudinal Radiology}
& B4, M, R-L, C; finding-level F1, change accuracy
& Medical change captioning. We report standard captioning metrics and an adapted finding-level evaluation with micro/macro F1 and change accuracy. \\

\textbf{Fine-grained Image Comparison}
& Base win, Reference win, tied, SVE win
& Reference-free MLLM-as-a-judge evaluation. The judge compares SVE against the stateless baseline and the reference editing instruction using pairwise preference counts. \\

\textbf{Remote Sensing}
& B4, M, R-L, C, $S_m^*$
& Multi-reference remote-sensing change captioning. $S_m^*$ denotes the average over BLEU-4, METEOR, ROUGE-L, and CIDEr. \\
\bottomrule
\end{tabularx}
\caption{
\textbf{Evaluation protocol by task.}
We report the metric sets used in the main result tables: regression errors for spatial aggregation, captioning and accuracy metrics for CLEVR, perplexity for VisGym behavioral cloning, captioning and finding-level metrics for radiology, pairwise judge preferences for ImgEdit, and multi-reference captioning metrics for LEVIR-CC.
}
\label{tab:metric_protocol_by_task}
\end{table*}

\subsection{Caption-metric conventions}

For captioning tasks, we distinguish between two metric conventions. The first is a lightweight sanity-check protocol based on common Python metric implementations. The second is the paper-aligned image-captioning protocol used by most prior work.

In the paper-aligned protocol, captions are first processed with standard caption-tokenization conventions before computing BLEU, METEOR, ROUGE-L, CIDEr, and SPICE. This matters because tokenizer choice and METEOR implementation can shift absolute values, especially on short or templated text. For comparisons with prior published results, we cite the paper-aligned protocol. The lightweight protocol is used only as an auxiliary view and for metrics such as BERTScore and perplexity that are not part of the standard captioning suite.

For perplexity, we compute token-weighted PPL under the same supervision mask used during training:
\[
\mathrm{PPL}
=
\exp\left(
\frac{\sum_i \mathrm{NLL}_i \cdot n_i}{\sum_i n_i}
\right),
\]
where $n_i$ is the number of supervised tokens for sample $i$. For single-shot captioning datasets, this means only the final answer tokens are included. For Visual Trajectory Behavioral Cloning, all action-generating assistant turns are supervised and included.

\subsection{CLEVR Multi-Change scoring}

Multi-object Visual Differencing requires a specialized scoring protocol because the reference caption describes multiple simultaneous changes whose sentence order carries no semantic content. A model should receive the same credit whether it lists the correct changes in the original order or in a different order.

We therefore score this task as permutation-invariant per-change captioning. First, each prediction and reference is split into individual change sentences. Each reference change is assigned a change type, such as addition, deletion, movement, or replacement. For each reference change, we construct a small set of valid lexical variants corresponding to the same underlying change. This accounts for the fact that the same edit can be described by several equivalent templates, such as ``a new object is visible'' and ``an object has been added.''

Next, we compute a pairwise similarity matrix between predicted change sentences and reference changes. The score for each pair is the best similarity between the predicted sentence and the allowed reference variants for that change. We then use one-to-one bipartite matching to find the assignment that maximizes total similarity. This makes the score invariant to the order in which changes are described.

\section{Training Configuration, Environment, and Infrastructure}
\label{app:training_infrastructure}

This appendix complements the data-formatting appendix (\cref{app:training_data_formatting}) with the software environment, distributed-training setup, shared hyperparameters, task-specific training choices, evaluation infrastructure, and hardware used in our experiments.

\subsection{Software environment}

\begin{formatbox}{Software stack}
All experiments use a single Python 3.12 environment with PyTorch, Transformers, Accelerate, FlashAttention, LlamaFactory, and standard captioning-evaluation libraries. Training uses bf16 precision throughout.
\end{formatbox}

\begin{table}[ht]
\centering
\small
\setlength{\tabcolsep}{5pt}
\renewcommand{\arraystretch}{1.08}
\resizebox{\linewidth}{!}{%
\begin{tabular}{lll}
\toprule
Component & Version / setting & Role \\
\midrule
Python & 3.12.13 & Runtime environment \\
PyTorch & 2.10.0 + CUDA 12.8 & Training backend \\
Transformers & 5.2.0 & Model implementation \\
Accelerate & 1.11.0 & Distributed training support \\
FlashAttention & 2.8.3 & Efficient attention kernels \\
LlamaFactory & local editable checkout & SFT framework \\
pycocoevalcap & latest available & Captioning metrics \\
Anthropic SDK & 0.102.0 & VLM-judge evaluation \\
Weights \& Biases & latest available & Training logs \\
\bottomrule
\end{tabular}%
}
\vspace{1ex}
\caption{
\textbf{Software environment.}
We use a fixed Python environment with PyTorch, Transformers, LlamaFactory, FlashAttention, and standard captioning-evaluation libraries.
}
\label{tab:software_environment}
\end{table}

\begin{table*}[t]
\centering
\caption{Default SFT hyperparameters used for each training setup. Batch layout follows per-device batch size, accumulation steps, and number of ranks.}
\label{tab:sft_hyperparams_all}
\small
\setlength{\tabcolsep}{3.2pt}
\renewcommand{\arraystretch}{1.08}
\begin{tabular}{lcccccc}
\toprule
\shortstack{\textbf{SFT}\\\textbf{Hyperparam.}}
& \shortstack{\textbf{Spatial}\\\textbf{Aggr.}}
& \shortstack{\textbf{Visual}\\\textbf{Diff.}}
& \shortstack{\textbf{Traj.}\\\textbf{Imit.}}
& \shortstack{\textbf{Long.}\\\textbf{Radiology}}
& \shortstack{\textbf{Image}\\\textbf{Comp.}}
& \shortstack{\textbf{Remote}\\\textbf{Sensing}} \\
\midrule
Base model
& \texttt{Qwen3.5-4B}
& \texttt{Qwen3.5-4B}
& \texttt{Qwen3.5-4B}
& \texttt{Qwen3.5-4B}
& \texttt{Qwen3.5-4B}
& \texttt{Qwen3.5-4B} \\

Global batch
& 384
& 384
& 384
& 384
& 384
& 384 \\

Batch layout
& $8{\times}6{\times}8$
& $4{\times}12{\times}8$
& $4{\times}12{\times}8$
& $8{\times}6{\times}8$
& $16{\times}3{\times}8$
& $8{\times}6{\times}8$ \\

Training length
& 500 steps
& 250 steps
& 250 steps
& 2 epochs
& 2 epochs
& 2 epochs \\

Learning rate
& $1.5{\times}10^{-5}$
& $1.5{\times}10^{-5}$
& $1.5{\times}10^{-5}$
& $1.5{\times}10^{-5}$
& $1.5{\times}10^{-5}$
& $2.0{\times}10^{-5}$ \\

LR scheduler
& Cosine
& Cosine
& Cosine
& Cosine
& Cosine
& Cosine \\

\texttt{mask\_history}
& \texttt{true}
& \texttt{true}
& \texttt{false}
& \texttt{true}
& \texttt{true}
& \texttt{true} \\

SVE $W_{o,2}$ std
& 0.0
& 0.0
& 0.0
& 0.0001
& 0.0001
& 0.0001 \\

Precision
& \texttt{bf16}
& \texttt{bf16}
& \texttt{bf16}
& \texttt{bf16}
& \texttt{bf16}
& \texttt{bf16} \\

Trainable modules
& Full
& Full
& Full
& Full
& Full
& Full \\

FSDP
& Full shard
& Full shard
& Full shard
& Full shard
& Full shard
& Full shard \\
\bottomrule
\end{tabular}
\end{table*}

\subsection{Distributed training}

All full-finetuning experiments use single-node, 8-GPU FSDP training. We use full-parameter finetuning rather than LoRA in all results.

\begin{formatbox}{FSDP setup}
\begin{itemize}
    \item \textbf{Parallelism.} Single-node FSDP with 8 H100 GPUs.
    \item \textbf{Sharding.} Full-shard FSDP over both the language-model decoder blocks and the vision-encoder blocks.
    \item \textbf{Checkpointing.} Gradient checkpointing is enabled with non-reentrant checkpointing for compatibility with FSDP.
\end{itemize}
\end{formatbox}

The vision tower is sharded together with the language model because our method modifies the visual encoder and trains it end-to-end. We also keep the language model, visual encoder, and multimodal projector trainable in all main experiments.

\subsection{Tokenized cache}

Training samples are tokenized and cached before training. The cache key includes the chat template, cutoff length, history-masking setting, and dataset identity. The same tokenized cache can be reused by the stateless baseline and SVE when the data-formatting settings match. This ensures that baseline and SVE runs consume identical text-image inputs.

\subsection{Task-specific training settings}

Refer to \cref{tab:sft_hyperparams_all} for default hyperparameters we use for each task in training.

\begin{formatbox}{Rationale for task-specific differences}
\begin{itemize}
    \item \textbf{Batch layout.} Per-device batch size is set to the largest value that fits in GPU memory.
    \item \textbf{Training duration.} Design-table tasks use short fixed-step training for fast ablations. Real-world tasks use epoch-based training.
    \item \textbf{History masking.} Real-world multi-image captioning tasks explicitly mask filler assistant turns; synthetic and agentic tasks either contain no filler or use every assistant turn as training signal.
    \item \textbf{SVE initialization.} Controlled design tasks use zero initialization, while real-world tasks use a tiny nonzero output-projection initialization ($\sigma=10^{-4}$, compared to a standard Transformer-scale $\sigma\approx 2\times10^{-2}$) after ablations showed better optimization.
\end{itemize}
\end{formatbox}

\section{Table view of different SVE designs}
\label{app:sve_different_view}

We provide a table view of different stateful visual encoder designs to complement \cref{fig:design_and_recipe}. This table additionally provides added parameters and compute for reference.
\begin{table*}[ht]
\centering
\small
\caption{
\textbf{Stateful visual encoder design space and per-layer overhead}.
Let $X \in \mathbb{R}^{N \times d}$ denote current-image tokens and $\past{Y} \in \mathbb{R}^{\past{K} \times d}$ denote predecessor-state tokens.
Each original visual encoder layer is abstracted as $\mathrm{FFN}_{\theta}(\mathrm{SA}_{\theta}(X))$.
\textsc{Self Ext.} reuses the pretrained self-attention module with an expanded attention mask, while \textsc{Cross} and \textsc{Cross+FFN} introduce separate state-conditioning modules.
\textsc{AdaLN-Zero} conditions the original block through pooled predecessor-state modulation.
Orange denotes predecessor-state information, and purple denotes newly initialized state-conditioning parameters/modules.
We report additions beyond the original stateless block, ignoring residual connections, normalization layers, positional embeddings, softmax costs and bias terms.
For parameter counts, one newly added attention module contains $W_Q,W_K,W_V,W_O \in \mathbb{R}^{d \times d}$, and one newly added FFN has shape $d \rightarrow d_{\mathrm{ff}} \rightarrow d$.
For \textsc{Self Ext.}, added compute counts only the extra current-to-predecessor score and value-attention terms induced by the expanded mask.
}
\vspace{1ex}
\label{tab:sve_design_cost}
\resizebox{\linewidth}{!}{
\begin{tabular}{llll}
\toprule
\textbf{Design}
& \textbf{Block form}
& \textbf{Added}
& \textbf{Added} \\
\textbf{}
& \textbf{}
& \textbf{params.}
& \textbf{compute} \\
\midrule
\textbf{Self-Ext.}
&
$\mathrm{FFN}_{\theta}\!\left(
    \mathrm{SA}_{\theta}(Q=X,KV=[X;\past{Y}])
\right)$
&
$0$
&
$2N\past{K}d$
\\
\midrule
\textbf{AdaLN-Zero}
&
$\begin{alignedat}{1}
&c = \mathrm{Pool}(\past{Y}),\quad
(\gamma_1,\beta_1,\alpha_1,\gamma_2,\beta_2,\alpha_2)=\newparam{g_{\phi}}(c),\\
&\alpha_2\odot \mathrm{FFN}_{\theta}\!\left((1+\gamma_2)\odot
\left[\alpha_1\odot \mathrm{SA}_{\theta}((1+\gamma_1)\odot X+\beta_1)\right]+\beta_2\right)
\end{alignedat}$
&
\newparam{$6d^2$}
&
\makecell[l]{$\past{K}d$ $+\,\newparam{6d^2}$ \\ $+\,6Nd$}
\\
\midrule
\textbf{Cross}
&
$\mathrm{FFN}_{\theta}\!\left(
    \mathrm{SA}_{\theta}\!\left(
        QKV=\newparam{\mathrm{CA}_{\phi}}(Q=X,KV=\past{Y})
    \right)
\right)$
&
\makecell[l]{\newparam{$4d^2$}}
&
\makecell[l]{$2(N+\past{K})d^2$ \\ $+\,2N\past{K}d$}
\\
\midrule
\textbf{Cross+FFN}
&
$\mathrm{FFN}_{\theta}\!\left(
    \mathrm{SA}_{\theta}\!\left(
        QKV=\newparam{\mathrm{FFN}_{\psi}}\!\left(
            \newparam{\mathrm{CA}_{\phi}}(Q=X,KV=\past{Y})
        \right)
    \right)
\right)$
&
\makecell[l]{\newparam{$4d^2$} \\ \newparam{$+\,2d d_{\mathrm{ff}}$}}
&
\makecell[l]{$2(N+\past{K})d^2$ \\ $+\,2N\past{K}d$ \\ $+\,2N d d_{\mathrm{ff}}$}
\\

\bottomrule
\end{tabular}
}
\end{table*}

\begin{table*}[th]
\centering\small
\caption{
\textbf{Finding categories used in evaluation.}
We list the 27 evaluated finding categories grouped by anatomy.
Numbers in parentheses indicate counts in the test set.
}
\label{tab:diffvqa_findings}
\setlength{\tabcolsep}{4pt}
\begin{tabular}{lp{0.70\linewidth}}
\toprule
Anatomy & Findings (test set count) \\
\midrule
Lungs                        & atelectasis (6{,}210), lung opacity (6{,}193), edema (3{,}499), pneumonia (3{,}257), consolidation (2{,}293), emphysema (616), infection (479), granuloma (140), contusion
(69) \\
Pleura                       & pleural effusion (5{,}075), pneumothorax (1{,}027), pleural thickening (419), blunting of the costophrenic angle (371) \\
Cardiac                     & cardiomegaly (3{,}671), vascular congestion (1{,}889), heart failure (283), hilar congestion (67) \\
Mediastinum / Aorta / Hernia & hernia (159), pneumomediastinum (98), tortuosity of the thoracic aorta (53), tortuosity of the descending aorta (5) \\
Chest wall / Skeletal        & calcification (833), fracture (746), scoliosis (202), hematoma (69) \\
Adjacent / Other             & air collection (56), gastric distention (11) \\
\bottomrule
\end{tabular}
\end{table*}

\section{Finding-level Medical-Diff-VQA Evaluation Details}
\label{app:diffvqa_eval}
In this section, we provide details of the evaluation pipeline of Medical-Diff-VQA results in \cref{tab:radiology_performance}.
\paragraph{Task setup.}
\label{app:diffvqa_task}
We evaluate VLMs on the Medical-Diff-VQA test split ($16{,}347$ paired chest X-rays).
During evaluation, we prompt VLMs with instruction
\emph{``This is the reference (prior) chest X-ray.
\dots This is the current chest X-ray. What has changed compared to the reference image?''}

\paragraph{Chest X-ray Finding Categories.}

We group the references of the Medical-Diff-VQA test into 27 finding categories as shown in~\cref{tab:diffvqa_findings}.

\paragraph{Parsing protocol.}
\label{app:diffvqa_parsing}

We describe how references and VLM free-form outputs are converted into a finding-level format for evaluation.
Specifically, we parse both references and model outputs with a regular-expression pipeline.
The references follow a templated structure with three direction categories: \emph{added}, \emph{missing}, and \emph{no change}.
For VLM outputs that do not follow the template, which account for less than 2\% of cases, the pipeline yields an empty tuple set, treating them as \textit{no change}.

\begin{center}
\resizebox{\linewidth}{!}{%
\begin{tabular}{@{}lll@{}}
\toprule
Direction & Matched phrase & Parsed tuple \\
\midrule
\emph{added}
& additional finding(s) of $\langle X_1, X_2, \ldots \rangle$
& $(X_i, \textit{added})$ \\
\emph{missing}
& missing the finding(s) of $\langle X_1, X_2, \ldots \rangle$
& $(X_i, \textit{missing})$ \\
\emph{no change}
& nothing has changed
& \emph{none} \\
\bottomrule
\end{tabular}%
}
\end{center}

\paragraph{Metric definitions.}

For each test pair $n \in \{1, \dots, N\}$ we form a $54$-dimensional binary gold vector $\mathbf{g}^{(n)} \in \{0,1\}^{54}$ indexed by the $27 \times 2$ (finding, direction) labels parsed from the reference,
and an analogous prediction vector $\mathbf{p}^{(n)}$ parsed from the model output. Per label $i$ we accumulate $\mathrm{TP}_i, \mathrm{FP}_i, \mathrm{FN}_i$ across all $N$ pairs and let
$\mathrm{F}_{1,i} = 2\mathrm{TP}_i / (2\mathrm{TP}_i + \mathrm{FP}_i + \mathrm{FN}_i)$. The metrics are calculated as follows:

\[
\resizebox{\linewidth}{!}{$
\begin{aligned}
\textbf{Micro F1}
&= \frac{2\, \sum_{i=1}^{54} \mathrm{TP}_i}
{2\, \sum_{i=1}^{54} \mathrm{TP}_i
+ \sum_{i=1}^{54} \mathrm{FP}_i
+ \sum_{i=1}^{54} \mathrm{FN}_i}, \\[2pt]
\textbf{Macro F1}
&= \frac{1}{54} \sum_{i=1}^{54} \mathrm{F}_{1,i}, \\[2pt]
\textbf{Change Acc.}
&= \frac{1}{N} \sum_{n=1}^{N}
\mathbb{1}\!\left[
\mathbf{g}^{(n)} = \mathbf{0}
\Leftrightarrow
\mathbf{p}^{(n)} = \mathbf{0}
\right].
\end{aligned}
$}
\]

No-change pairs do not contain finding-level annotations in the reference: the label is \emph{``nothing has changed''}.
We therefore represent these cases with an all-zero finding vector, indicating that no added or missing findings are present.
We calculate finding-level F1 on the 14{,}030 pairs whose references identify at least one specific change.
\emph{Change Accuracy} reports the complementary pair-level binary metric: whether the model correctly recognizes that the patient is stable.

\paragraph{Per-anatomy breakdown}
Table~\ref{tab:diffvqa_percategory} decomposes the Micro F1 by the anatomical grouping of Table~\ref{tab:diffvqa_findings} to provide detailed analysis of SVE versus stateless baseline.

\begin{table}[t]
\centering
\small
\caption{Per-anatomy Micro F1 of finding-level evaluation under greedy decoding.}
\label{tab:diffvqa_percategory}
\setlength{\tabcolsep}{4pt}
\renewcommand{\arraystretch}{1.08}
\begin{tabular}{lrrrr}
\toprule
Anatomy & \# findings & Stateless & SVE & $\Delta$ \\
\midrule
Lungs                         & 9 & 31.56 & 32.17 & $+0.61$ \\
Pleura                        & 4 & 41.10 & 42.03 & $+0.93$ \\
Cardiac                       & 4 & 24.72 & 25.42 & $+0.70$ \\
\begin{tabular}[t]{@{}l@{}}
Mediastinum / Aorta \\
/ Hernia
\end{tabular}
                              & 4 &  7.18 & 12.32 & $+5.13$ \\
Chest wall / Skeletal         & 4 &  8.89 &  8.51 & $-0.37$ \\
Adjacent / Other              & 2 &  0.00 &  0.00 & $\phantom{+}0.00$ \\
\bottomrule
\end{tabular}
\end{table}

\section{AI Use Disclosure}
\label{app:ai}

The authors used AI-based tools to assist with code generation, editing, and writing during the preparation of this paper. Specifically, AI assistance was used to help draft and revise portions of the manuscript for clarity, grammar, and organization, and to support the development, debugging, and refinement of code used in the research workflow. All AI-generated or AI-assisted content, code, analyses, and interpretations were reviewed, verified, and, where necessary, modified by the authors. The authors take full responsibility for the accuracy, integrity, originality, and final content of the paper, including any code or text developed with AI assistance.

\end{document}